\documentclass[letterpaper]{article} 
\usepackage{aaai24}  
\usepackage{times}  
\usepackage{helvet}  
\usepackage{courier}  
\usepackage[hyphens]{url}  
\usepackage{graphicx} 
\usepackage{natbib}  
\usepackage{caption} 
\usepackage{algorithm}
\usepackage{algorithmic}

\usepackage{newfloat}
\usepackage{listings}
\DeclareCaptionStyle{ruled}{labelfont=normalfont,labelsep=colon,strut=off} 
\floatstyle{ruled}
\newfloat{listing}{tb}{lst}{}
\floatname{listing}{Listing}
\usepackage[table]{xcolor}
\usepackage{amssymb}
\usepackage{booktabs}
\usepackage{caption}
\usepackage{subcaption}
\usepackage{pifont}
\usepackage{makecell}
\usepackage{multirow}
\usepackage{arydshln}
\usepackage{listings}
\makeatletter
\@namedef{ver@everyshi.sty}{}
\makeatother
\usepackage{nicematrix}

\definecolor{bgpink}{RGB}{254, 238, 237}

\newcommand{\LN}{\texttt{LN}}
\newcommand{\starrelu}{\texttt{StarReLU}}

\newcommand{\AFNO}{\texttt{AFNO}}
\newcommand{\pwconva}{\texttt{Conv}_\textrm{pw1}}
\newcommand{\pwconvb}{\texttt{Conv}_\textrm{pw2}}
\newcommand{\X}{\mathbf{X}}

\title{FFT-based Dynamic Token Mixer for Vision}
\author{
    Yuki Tatsunami\textsuperscript{\rm 1 2}
    Masato Taki\textsuperscript{\rm 1}
}
\affiliations{
    \textsuperscript{\rm 1}Rikkyo University\\
    \textsuperscript{\rm 2}AnyTech Co., Ltd.\\
    y.tatsunami@rikkyo.ac.jp, taki\_m@rikkyo.ac.jp
}

\begin{document}

\maketitle

\begin{abstract}
Multi-head-self-attention (MHSA)-equipped models have achieved notable performance in computer vision. Their computational complexity is proportional to quadratic numbers of pixels in input feature maps, resulting in slow processing, especially when dealing with high-resolution images. New types of token-mixer are proposed as an alternative to MHSA to circumvent this problem: an FFT-based token-mixer involves global operations similar to MHSA but with lower computational complexity. However, despite its attractive properties, the FFT-based token-mixer has not been carefully examined in terms of its compatibility with the rapidly evolving MetaFormer architecture. Here, we propose a novel token-mixer called Dynamic Filter and novel image recognition models, DFFormer and CDFFormer, to close the gaps above. The results of image classification and downstream tasks, analysis, and visualization show that our models are helpful. Notably, their throughput and memory efficiency when dealing with high-resolution image recognition is remarkable. Our results indicate that Dynamic Filter is one of the token-mixer options that should be seriously considered. The code is available at https://github.com/okojoalg/dfformer
\end{abstract}

\section{Introduction}
A transformer architecture was propelled to the forefront of investigations in the computer vision field. The architecture locates the center in various visual recognition tasks, including not only image classification \cite{dosovitskiy2020image, touvron2020training, yuan2021tokens, zhou2021refiner, wang2021pyramid, beyer2023flexivit} but also action recognition \cite{neimark2021video, bertasius2021space}, even point cloud understanding \cite{guo2021pct, zhao2021point, wei2022spatial}. Vision Transformer (ViT) \cite{dosovitskiy2020image} and its variants triggered this explosion. ViT was inspired by Transformer in NLP and is equipped with a multi-head self-attention (MHSA) mechanism \cite{vaswani2017attention} as a critical ingredient. MHSA modules are low-pass filters \cite{park2022vision}. Hence they are suitable for recognizing information about an entire image. While MHSA modules have been a success, it has faced challenges, especially in aspects of quadratic computational complexity due to global attention design. This problem is not agonizing in ImageNet classification but in dense tasks like semantic segmentation since we often deal with high-resolution input images. This problem can be tackled by using local attention design \cite{liu2021swin, zhang2021multi, chu2021twins, chen2021regionvit}, but the token-to-token interaction is limited, which means that one of the selling points of Transformers is mislaid.

\begin{figure}[tb]
\centering
\begin{minipage}[t]{0.45\hsize}
\centering
\includegraphics[width=1.\linewidth]{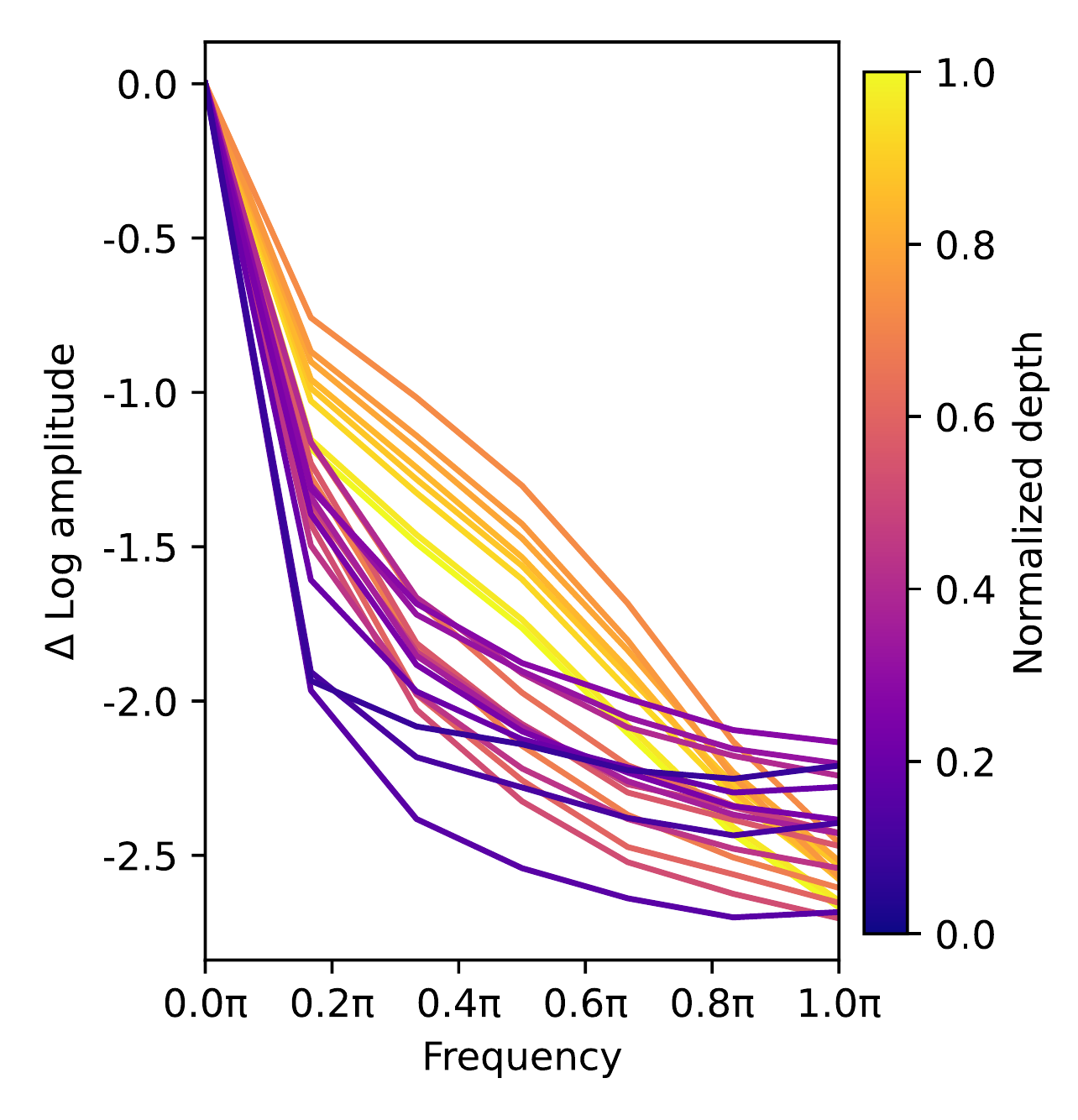}\subcaption{\label{fig:logamp_vit}ViT}
\end{minipage}
\begin{minipage}[t]{0.45\hsize}
\centering
\includegraphics[width=1.\linewidth]{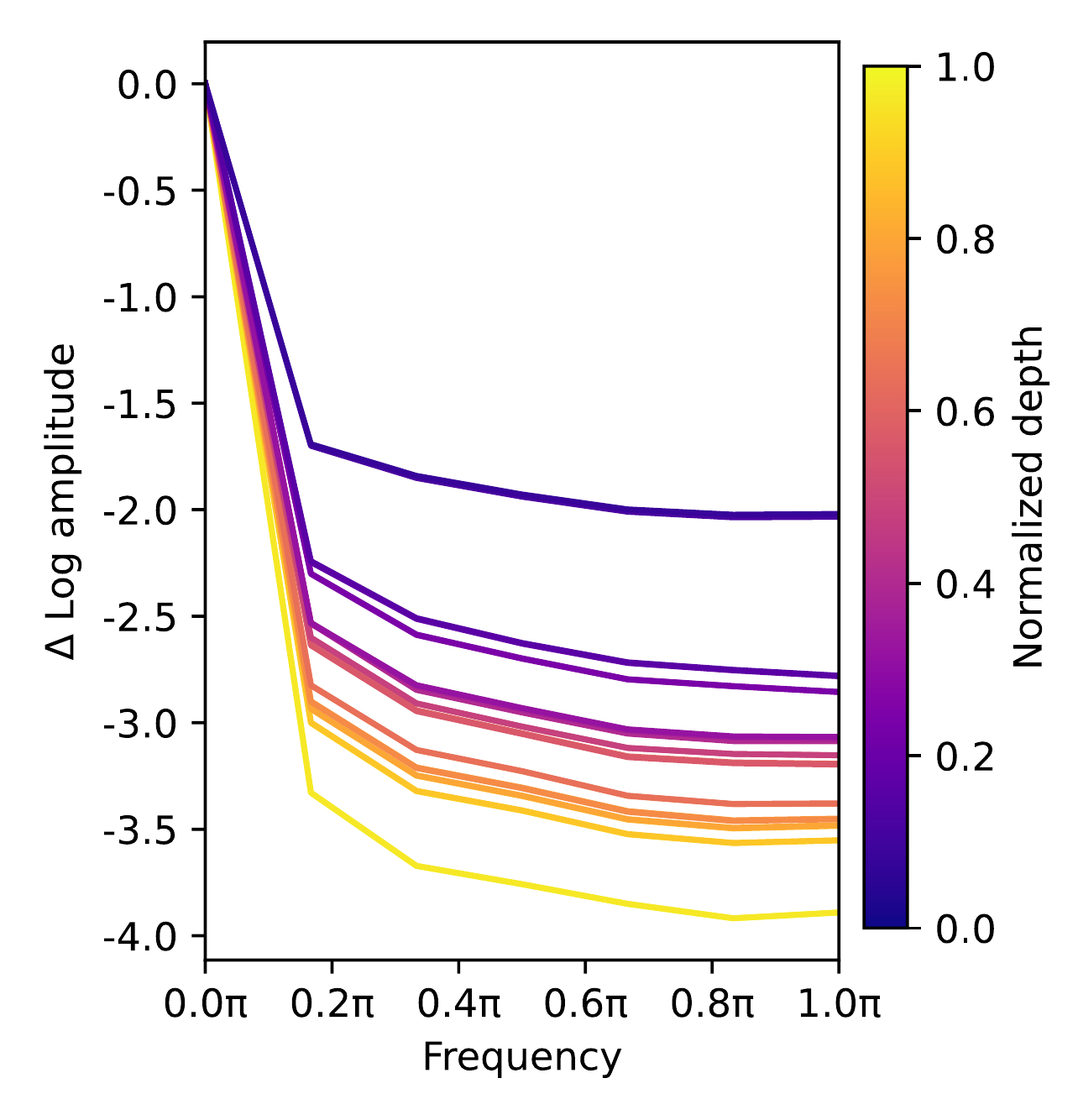}
\subcaption{\label{fig:logamp_gfnet}GFNet}
\end{minipage}
\caption{\label{fig:logamp}Relative log amplitudes of Fourier transformed feature maps. $\Delta$ Log amplitude means relative logarithmic amplitude concerning the logarithmic amplitude at normalized frequency 0. The yellower the color, the deeper the layer.}
\end{figure}

GFNet \cite{rao2021global} is one of the architectures induced by ViT and has two charming properties: (1) Global filter, the principal component in GFNet, multiplies features and complex parameters in frequency domains to increase/decrease a specific frequency. It is similar to a large kernel convolution \cite{ding2022scaling, liu2022more} but has the attractive aspect of reducing theoretical computational costs \cite{rao2021global}. In addition, GFNet has less difficulty training than a large kernel convolution, whereas large kernel convolutions employ a tricky training plan. Moreover, a computational complexity of GFNet is $\mathcal{O}(HWC[\log_2(HW)]+HWC)$, which is superior to of self-attention $\mathcal{O}(HWC^2 + (HW)^2C)$ where $H$ is height, $W$ is width, and $C$ is channel. In other words, the higher the resolution of the input image, the more complexity GFNet has relative dominance. (2) A global filter is expected to serve as a low-pass filter like MHSA because a global filter is a global operation with room to capture low-frequency information. We follow \cite{park2022vision} and compare the relative log amplitudes of Fourier transformed between ViT, and GFNet, which is a non-hierarchical architecture similar to ViT, and show in Figure~\ref{fig:logamp}. The result demonstrates that GFNet retains the same low-frequency information near 0 as ViT. Thus, we need to address the potential for the global filter alternative to MHSA due to its computational advantage when handling high-resolution input images and its similarity to MHSA.
Global filters and MHSA are different in certain respects, even though they have low-pass filter characteristics like they are different in certain respects. For example, MHSA is a data-dependent operation generated from data and applied to data. In contrast, global filters multiply data and parameters and hence are less data-dependent operations. In recent works, MLP-based models, which have been less data-dependent, employ data-dependent operations \cite{wang2022dynamixer, rao2022amixer}, achieved accuracy improvements. It is inferred from the findings that introducing a data-dependent concept to global filters will improve accuracy.

Meanwhile, one should note that global filters have not been based on the most up-to-date and effective architectures relative to the well-researched MHSA-based and their relatives' architectures. MetaFormer \cite{yu2021metaformer, yu2022metaformer} pays attention to the overall architecture of the Transformer. MetaFormer has a main block consisting of an arbitrary token-mixer and a channel mixer. It also makes sense to consider this framework for global filters because it is one of the most sophisticated architectures. It is necessary to close these differentials of architectural design if global filters can be a worthy alternative to MHSA modules.

We propose DFFormer, an FFT-based network to fill the above gap. Our architectures inherit MetaFormer and equip modules that can dynamically generate a global filter for each pair of feature channels in the image, considering their contents. Our experimental results show that the proposed methods achieve promising performance among vision models on the ImageNet-1K dataset, except for MHSA. DFFormer-B36 with 115M parameters has achieved a top-1 accuracy of 84.8\%. CDFFormer, a hybrid architecture with convolutions and FFTs architecture, is even better. Moreover, when dealing with high-resolution images, the proposed method realizes higher throughput than the MHSA-based architecture and architecture using both CNN and MHSA. We also found that the dynamic filter tends to learn low frequency rather than MHSA on a hierarchical Metaformer.

Our main contributions are summarized as follows: First, we have introduced the dynamic filter, a dynamic version of the traditional global filter. Second, we found that injecting Metaformer-like architecture using the dynamic filter can close the gap between the accuracy of the global filter architecture and of SOTA models. Finally, through comprehensive analysis, we have demonstrated that the proposed architecture is less costly for downstream tasks of higher resolution, like dense prediction.

\section{Related Work}
\paragraph{Vision Transformers and Metaformers}
Transformer \cite{vaswani2017attention}, which was proposed in NLP, has become a dominant architecture in computer vision as well, owing to the success of ViT \cite{dosovitskiy2020image} and DETR \cite{carion2020end}. Soon after, MLP-Mixer \cite{tolstikhin2021mlp} demonstrates that MLP can also replace MHSA in Transformer. MetaFormer \cite{yu2021metaformer}, as an abstract class of Transformer and MLP-Mixer, was proposed as a macro-architecture. The authors tested the hypothesis using MetaFormer with pooling. Generally, the abstract module corresponding to MHSA and MLP is called a token-mixer. It emphasizes the importance of MetaFormer that new classes of MetaFormer, such as Sequencer \cite{tatsunami2022sequencer} using RNNs, Vision GNN \cite{han2022vision} using graph neural networks, and RMT \cite{fan2023rmt} using retention mechanism that is variant of the attention mechanism, have emerged. Moreover, a follow-up study \cite{yu2022metaformer} has shown that even more highly accurate models can be developed by improving activation and hybridizing with multiple types of tokens. We retain the macro-architecture of \cite{yu2022metaformer}. The FFT-based module is a global operation like MHSA and can efficiently process high-resolution images without much loss of accuracy.
\paragraph{FFT-based Networks} In recent years, neural networks using Fourier transforms have been proposed. FNet \cite{lee2021fnet}, designed for NLP, contains modules using a discrete or fast Fourier Transform to extract only the real part. Accordingly, they have no parameters. Fast-FNet \cite{sevim2023fastfnet} removes waste and streamlines FNet. GFNet \cite{rao2021global} is an FFT-based network designed for vision with global filters. The global filter operates in frequency space by multiplying features with a Fourier filter, equivalent to a cyclic convolution. The filter is a parameter, i.e., the same filter is used for all samples. In contrast, our dynamic filter can dynamically generate the Fourier filter using MLP. \cite{guibas2021adaptive} is one of the most related works and proposed AFNO. Instead of the element-wise product, the MLP operation of complex weights is used to represent an adaptive and non-separable convolution. AFNO is not a separable module. As a result, the computational cost is higher than the separable module. On the contrary, our proposed method can realize a dynamic separatable Fourier filter, and our models do not differ much from the global filter in terms of FLOPs. We will mention these throughputs in subsection~\ref{ssec:ablation}. Concurrent work \cite{vodynamic} also attempts to generate dynamic filters, but the structure is such that only the filter coefficients are changed. Although the amplitude and frequency of the filter are dynamic, the major property of the filter cannot be changed (e.g., a high-pass filter cannot be changed to a low-pass filter). In contrast, a dynamic filter can represent high-pass and low-pass filters by first-order coupling.

\paragraph{Dynamic Weights} There have been some suggestive studies on the generation of dynamic weights. \cite{jia2016dynamic} realizes the filter parameters to change dynamically, where the filter weights of the convolution were model parameters. A filter generation network generates the filters. \cite{ha2016hypernetworks} is a contemporaneous study. It
proposed HyperNetworks, neural networks to create weights of another neural network. They can generate weights of CNN and RNN. \cite{yang2019condconv,chen2020dynamic} are at the same time worked on, both of which make the filters of convolution dynamic. These works influence our dynamic filter: whereas \cite{yang2019condconv, chen2020dynamic} predict the real coefficients of linear combination for a real filter basis of standard convolution, our work predicts the real coefficients of linear combination for a complex parameter filter basis. This method can be considered an equivalent operation in forward propagation from the fast Fourier transform's linearity; thus, our work is similar to these works. In backpropagation, however, whether the proposed dynamic filter can be learned is nontrivial. Our architecture does not also need the training difficulties cited in these studies, such as restrictions on batch size or adjustment of softmax temperatures. Outside of convolution, Synthesizer \cite{tay2021synthesizer}, DynaMixer \cite{wang2022dynamixer}, and AMixer \cite{rao2022amixer} have MLP-Mixer-like token-mixer, but with dynamic generation of weights. In particular, AMixer includes linear combinations related to our work.
\section{Method}
\subsection{Preliminary: Global Filter}
We will look back at a discrete Fourier transform before introducing a global filter \cite{rao2021global}. We discuss a 2D discrete Fourier transform (2D-DFT). For given the 2D signal $x(h,w)$, we define the 2D-DFT $\tilde{x}(h',w')$ as following:
\begin{align}
    \tilde{x}(h',w') &= \sum_{h=0}^{H-1}\sum_{w=0}^{W-1} \frac{x(h,w) e^{-2\pi j\left(\frac{hh'}{H}+\frac{ww'}{W}\right)}}{\sqrt{HW}}
\end{align}
where $H, W \in \mathbb{N}$, $h, h' \in \{z \in \mathbb{Z}\, |\, 0 < z < H \}$, and $w, w' \in \{z \in \mathbb{Z}\, |\, 0 < z < W \}$. Its inverse transformation exists and is well-known as a 2D inverse discrete Fourier transform (2D-IDFT). Generally, $\tilde{x}(h',w')$ is complex and periodic to $h'$ and $w'$. We assume that $x(h',w')$ is a real number, then a complex matrix $\tilde{X}(h',w')$ associated with $\tilde{x}(h',w')$ is Hermitian. A space to which $\tilde{x}(h',w')$ belongs is known as a frequency domain and is available for analyzing frequencies. In addition, the frequency domain has a significant property: Multiplication in the frequency domain is equivalent to a cyclic convolution in the original domain, called convolution theorem \cite{proakis2007digital}. The 2D-DFT is impressive but has a complexity of $\mathcal{O}(H^2W^2)$. Therefore, a 2D-FFT is proposed and is often used in signal processing. It is improved with complexity $\mathcal{O}(HW\log_2{(HW)})$.

Second, we define the global filter for the feature $\X \in \mathbb{R}^{C\times H \times W}$. The global filter $\mathcal{G}$ formulate the following:
\begin{align}
    \mathcal{G}(\X) &= \mathcal{F}^{-1}(\mathbf{K} \odot \mathcal{F}(\X))
\end{align}
where $\odot$ is the element-wise product, $\mathbf{K} \in \mathbb{C}^{C\times H \times \lceil \frac{W}{2}\rceil}$ is a learnable filter, and $\mathcal{F}$ is the 2D-FFT of which redundant components are reduced (i.e., \texttt{rfft2}) since $\mathcal{F}(x)$ is Hermitian. Note that the operation is equivalent to cyclic convolution of the filter $\mathcal{F}^{-1}(\X)$ based on the convolution theorem.

The global filter is known to have some properties. (1) The theoretical complexity is $\mathcal{O}(HWC\lceil\log_2{(HW)}\rceil+HWC)$. It is more favorable than Transformer and MLP when input is high resolution. (2) The global filter can easily scale up the input resolution owing to interpolating the filter. See \cite{rao2021global} for more details.
\subsection{Dynamic Filter}
\begin{figure}[tb]
\begin{minipage}[b]{1.\hsize}
  \centering
  \includegraphics[width=1.\linewidth]{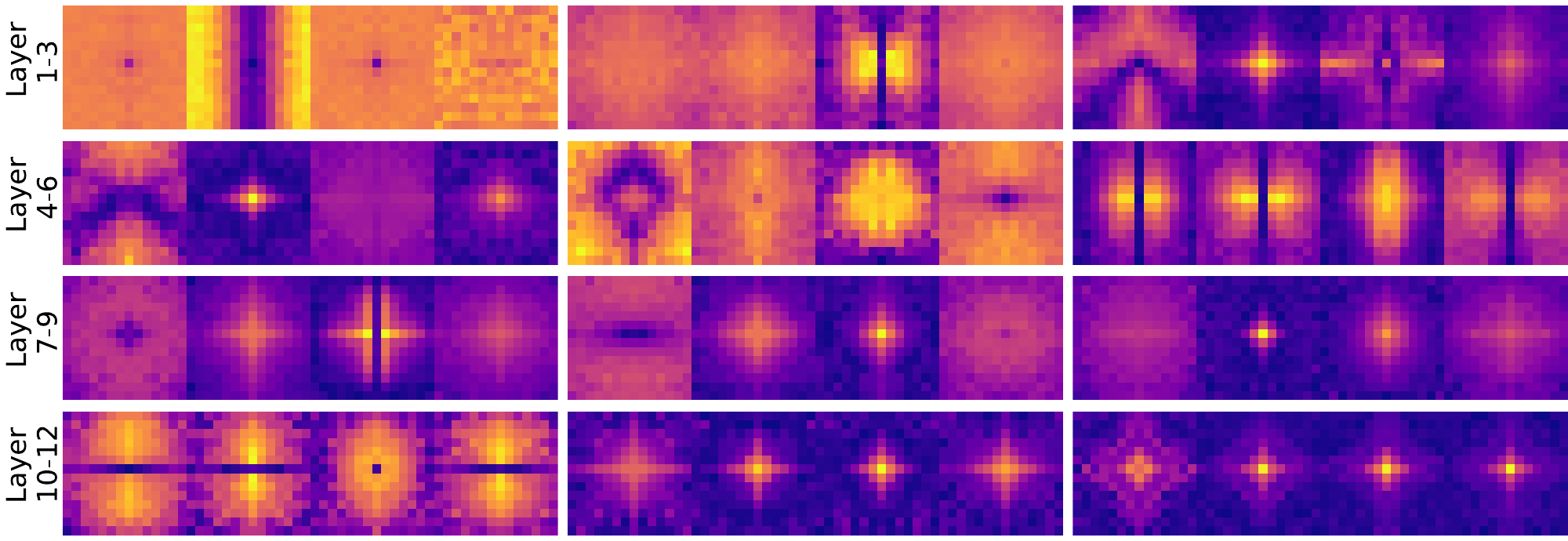}
  \caption{\label{fig:gfnet_ti_filter}Filters in the frequency domain on GFNet-Ti}
\end{minipage}
\end{figure}
\begin{figure}[tb]
\centering
\includegraphics[width=.85\linewidth]{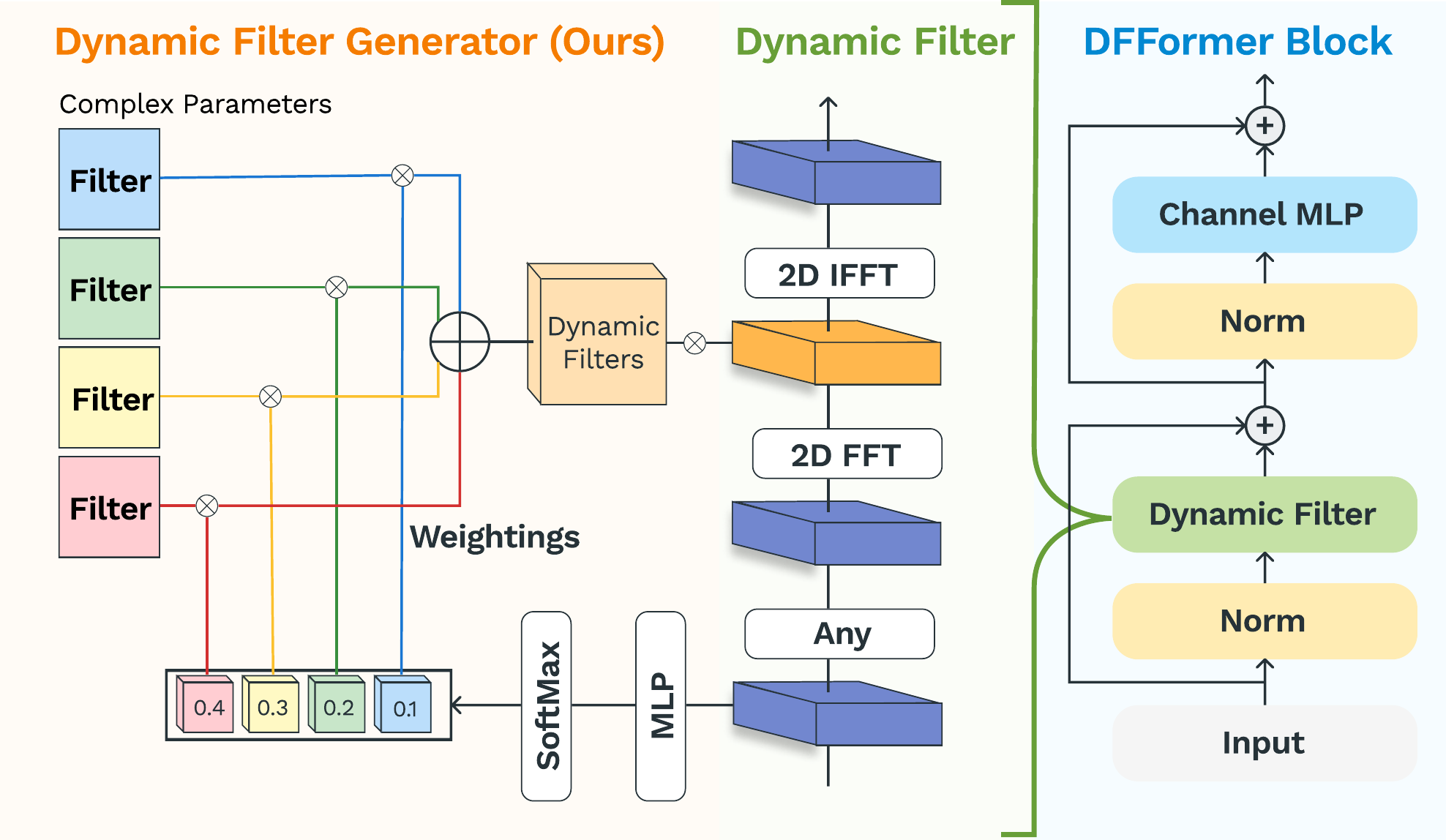}
\caption{\label{fig:modules} \textbf{Dynamic Filter} is our proposed component where "Any" modules are allowed continuous real maps. \textbf{DFFormer Block} is a MetaFormer block equipped with the dynamic filter.}
\end{figure}
This subsection discusses a dynamic filter in which a neural network dynamically determines an adequate global filter. The left side of Figure~\ref{fig:modules} shows the dynamic filter, which has a global filter basis of which the dimension is $N$, and linearly coupled global filters for them are used for each channel instead of learnable global filters. The coefficients are ruled by an MLP $\mathcal{M}$. We specify the whole dynamic filter $\mathcal{D}$ as follows:
\begin{align}
    \mathcal{D}(\X) &= \mathcal{F}^{-1}(\mathcal{K}_\mathcal{M}(\X) \odot \mathcal{F}\circ\mathcal{A}(\X))
\end{align}
where $\mathcal{K}_\mathcal{M}$ denotes the function that determines the dynamic filter. $\mathcal{A}$ are continuous real maps, including point-wise convolutions and identity maps.
\paragraph{Generating Dynamic Filter} We denote global filter basis $\mathbb{K}$ so that $\mathbb{K}=\{\mathcal{K}_1,\ldots,\mathcal{K}_N\}$ and $\mathcal{K}_1,\ldots,\mathcal{K}_N \in \mathbb{C}^{H \times \lceil \frac{W}{2}\rceil}$. Filters $\mathcal{K}_\mathcal{M}(\X)\in \mathbb{C}^{C'\times H \times \lceil \frac{W}{2}\rceil}$ are associated with $\mathcal{M}$ for weighting and are defined by the following:
\begin{align}
    \mathcal{K}_\mathcal{M}(\X)_{c,:,:}:=\sum_{i=1}^N \left(\dfrac{e^{s_{(c-1)N+i}}}{\sum_{n=1}^N{e^{s_{(c-1)N+n}}}}\right)\mathcal{K}_i, \\
    {\rm where}\; (s_{1}, \ldots, s_{NC'})^\top = \mathcal{M}\left(\dfrac{\sum_{h,w}{\X_{:,h,w}}}{HW}\right).
\end{align}
In this paper, we use $N=4$, which is the dimension of $\mathbb{K}$, to avoid over-computing. See ablation in subsection~\ref{ssec:ablation}.
\paragraph{MLP for weighting} We describe MLP $\mathcal{M}$ for weighting throughout specific calculation formulas:
\begin{equation}
    \mathcal{M}(\X) = W_2\starrelu(W_1\LN(\X)),
\end{equation}
where $\LN(\cdot)$ is layer normalization \cite{ba2016layer}, $\starrelu(\cdot)$ is an activation function proposed by \cite{yu2022metaformer}, $W_1 \in \mathbb{R}^{C\times {\tt int}(\rho C)}$, $W_2 \in \mathbb{R}^{{\tt int}(\rho C)\times NC'}$ denote matrices of MLP layer, $\rho$ is a ratio of the intermediate dimension to the input dimension of MLP. We choose $\rho=0.25$ but see section~\ref{ssec:ablation} for the other case.
\subsection{DFFormer and CDFFormer}
\begin{figure}[tb]
  \centering
  \includegraphics[width=.9\linewidth]{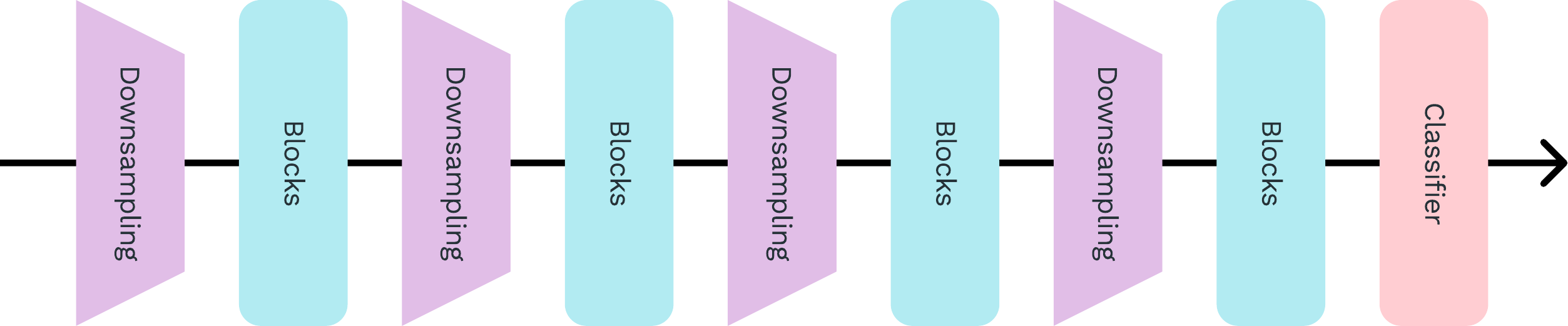}
  \caption{\label{fig:four_stage} A four-stage model}
\end{figure}
\begin{table}[tb]
    \centering
    \begin{tabular}{rr|cc}
\multicolumn{2}{r}{Model} & DFFormer & CDFFormer \\
\midrule
\multicolumn{2}{r|}{TokenMixer} & $\texttt{DF},\texttt{DF},\texttt{DF},\texttt{DF}$ & $\texttt{CF},\texttt{CF},\texttt{DF},\texttt{DF}$ \\ \hline
\multicolumn{2}{r|}{\multirow{2}{*}{Down Smp.}} & \multicolumn{2}{c}{\multirow{2}{*}{\makecell[c]{$K=7\text{-}3\text{-}3\text{-}3$, $S=4\text{-}2\text{-}2\text{-}2$, \\ $P=2\text{-}1\text{-}1\text{-}1$}}} \\
\multicolumn{2}{c|}{} & \multicolumn{2}{c}{} \\ \hline
\multirow{4}{*}{Size} & S18 & \multicolumn{2}{c}{$L=3\text{-}3\text{-}9\text{-}3$, $C=64\text{-}128\text{-}320\text{-}512$} \\
 & S36 & \multicolumn{2}{c}{$L=3\text{-}3\text{-}9\text{-}3$, $C=64\text{-}128\text{-}320\text{-}512$} \\
 & M36 & \multicolumn{2}{c}{$L=3\text{-}3\text{-}9\text{-}3$, $C=96\text{-} 192\text{-}384\text{-}576$} \\
 & B36 & \multicolumn{2}{c}{$L=3\text{-}3\text{-}9\text{-}3$, $C=128\text{-}256\text{-}512\text{-}768$} \\ \hline
 \multicolumn{2}{r|}{Classifier} &  \multicolumn{2}{c}{GAP, Layer Norm., MLP} \\
\bottomrule
\end{tabular}    \caption{\label{tab:overall}\textbf{Model settings of DFFormer and CDFFormer.} \texttt{DF} and \texttt{CF} mean DFFormer block and CFFormer block, respectively. $K$, $S$, $P$, $C$, and $L$ denote kernel size, stride, padding, number of channels, and number of blocks, respectively. The indices of each tuple correspond to the order of the stages.}
\end{table}
We construct DFFormer and CDFFormer complying with MetaFormer \cite{yu2022metaformer}. DFFormer and CDFFormer mainly consist of MetaFormer blocks, such as DFFormer and ConvFormer blocks. DFFormer and CDFFormer blocks comply with MetaFormer blocks
\begin{equation}
    \mathcal{T}(\X) = \X+\pwconvb\circ\mathcal{L}\circ\starrelu\circ\pwconva\circ\LN(\X)
\end{equation}
where $\LN(\cdot)$ is Layer Normalization \cite{ba2016layer}, $\pwconva(\cdot)$ and $\pwconvb(\cdot)$ are point-wise convolutions so that the number of output channels is $(C'=)2C$ and $C$, respectively, and $\mathcal{L}(\cdot)$ is dimensional invariant any function. If $\mathcal{L}(\cdot)$ is separable convolution, $\mathcal{T}$ is named as ConvFormer block at \cite{yu2022metaformer}. If $\mathcal{L}(\cdot)=\mathcal{F}^{-1}(\mathcal{K}_\mathcal{M}(\X) \odot \mathcal{F}(\cdot))$, we define that $\mathcal{T}$ is DFFormer block. A schematic diagram is shown on the right side of Figure~\ref{fig:modules} in order to understand the DFFormer blocks' structure. See codes for details in the appendix.

The overall framework is also following MetaFormer \cite{yu2021metaformer, yu2022metaformer}. In other words, we utilize the four-stage model in Figure~\ref{fig:four_stage}. We prepared four sizes of models for DFFormer, which mainly consists of DFFormer blocks, and CDFFormer, which is a hybrid model of DFFormer blocks and ConvFormer blocks. Each model is equipped with an MLP classifier with Squared ReLU~\cite{so2021primer} as the activation. A detailed model structure is shown in Table~\ref{tab:overall}. 
\section{Experiments}
We conduct experiments on ImageNet-1K benchmark \cite{krizhevsky2012imagenet} and perform further experiments to confirm downstream tasks such as ADE20K \cite{zhou2017scene}. Finally, we will conduct ablation studies about design elements.
\subsection{Image Classification}\label{ssec:classification}
\begin{table}[!bth]
    \centering
    \small

\begin{tabular}{lcrrrr}
\toprule
\multirow{2}{*}{\makecell[c]{Model}}   & \multirow{2}{*}{\makecell[c]{Ty.}}     & \multirow{2}{*}{\makecell[c]{Prm. \\ (M)}}   & \multirow{2}{*}{\makecell[c]{FLOP \\ (G)}}  & \multirow{2}{*}{\makecell[c]{Thrp. \\ (img/s)}} & \multirow{2}{*}{\makecell[c]{Top-1 \\ (\%)}} \\
~ & ~ & ~ & ~ & ~ & ~ \\
\midrule
ConvNeXt-T &C &29 &4.5 &1471 &82.1 \\
ConvFormer-S18 &C &27 &3.9 &756 &83.0 \\
CSWin-T &A &23 &4.3 &340 &82.7 \\
MViTv2-T &A &24 &4.7 &624 &82.3 \\
DiNAT-T & A & 28 & 4.3 & 816 & 82.7 \\
DaViT-Tiny & A & 28 & 4.5 & 1121 & 82.8 \\
GCViT-T & A & 28 & 4.7 & 566 & 83.5 \\
MaxViT-T &A &41 &5.6 &527 &83.6 \\
RMT-S & R & 27 & 4.5 & 406 & 84.1 \\
AMixer-T &M &26 &4.5 &724 &82.0 \\
GFNet-H-S &F &32 &4.6 &952 &81.5 \\
\textbf{DFFormer-S18}&F &30 &3.8 &535 &83.2 \\
CAFormer-S18 &CA &26 &4.1 &741 &83.6 \\
\textbf{CDFFormer-S18}&CF &30 &3.9 &567 &83.1 \\
\hline ConvNeXt-S &C &50 &8.7 &864 &83.1 \\
ConvFormer-S36 &C &40 &7.6 &398 &84.1 \\
MViTv2-S &A &35 &7.0 &416 &83.6 \\
DiNAT-S & A & 51 & 7.8 & 688 & 83.8 \\
DaViT-Small & A & 50 & 8.8 & 664 & 84.2 \\
GCViT-S & A & 51 & 8.5 & 478 & 84.3 \\
RMT-B & R & 54 & 9.7 & 264 & 85.0 \\
DynaMixer-S &M &26 &7.3 &448 &82.7 \\
AMixer-S &M &46 &9.0 &378 &83.5 \\
GFNet-H-B &F &54 &8.6 &612 &82.9 \\
\textbf{DFFormer-S36}&F &46 &7.4 &270 &84.3 \\
CAFormer-S36 &CA &39 &8.0 &382 &84.5 \\
\textbf{CDFFormer-S36}&CF &45 &7.5 &319 &84.2 \\
\hline ConvNeXt-B &C &89 &15.4 &687 &83.8 \\
ConvFormer-M36 &C &57 &12.8 &307 &84.5 \\
MViTv2-B &A &52 &10.2 &285 &84.4 \\
GCViT-S2 & A & 68 & 10.7 & 415 & 84.8 \\
MaxViT-S &A &69 &11.7 &449 &84.5 \\
DynaMixer-M &M &57 &17.0 &317 &83.7 \\
AMixer-B &M &83 &16.0 &325 &84.0 \\
\textbf{DFFormer-M36}&F &65 &12.5 &210 &84.6 \\
CAFormer-M36 &CA &56 &13.2 &297 &85.2 \\
\textbf{CDFFormer-M36}&CF &64 &12.7 &254 &84.8 \\
\hline ConvNeXt-L &C &198 &34.4 &431 &84.3 \\
ConvFormer-B36 &C &100 &22.6 &235 &84.8 \\
MViTv2-L &A &218 &42.1 &128 &85.3 \\
DiNAT-B & A & 90 & 13.7 & 499 & 84.4 \\
DaViT-Base & A & 88 & 15.5 & 528 & 84.6 \\
GCViT-B & A & 90 & 14.8 & 367 & 85.0 \\
MaxViT-B &A &120 &23.4 &224 &85.0 \\
RMT-L & R & 95 & 18.2 & 233 & 85.5 \\
DynaMixer-L &M &97 &27.4 &216 &84.3 \\
\textbf{DFFormer-B36}&F &115 &22.1 &161 &84.8 \\
CAFormer-B36 &CA &99 &23.2 &227 &85.5 \\
\textbf{CDFFormer-B36}&CF &113 &22.5 &195 &85.0 \\
\bottomrule
\end{tabular}

    \caption{\label{tab:imagenet}
    \textbf{Performance comparison of models pre-trained on ImageNet-1K at the resolution of $224^2$.} Throughput has been benchmarked on a V100 with 16GB memory at a batch size of 16. The full names of types are C is Convolution, A is Attention, R is Retention, M is MLP, F is FFT, CA is a hybrid of Convolution and Attention, and a hybrid of Convolution and FFT, respectively.}
\end{table}
DFFormers and CDFFormers have experimented on ImageNet-1K \cite{krizhevsky2012imagenet}, one of the most renowned data sets in computer vision for image classification. It has 1000 classes and contains 1,281,167 training images and 50,000 validation images. Our training strategy is mainly according to \cite{touvron2020training} and is detailed as follows. For data augmentation methods, we apply MixUp \cite{zhang2017mixup}, CutMix \cite{yun2019cutmix}, random erasing \cite{zhong2020random}, and RandAugment \cite{cubuk2020randaugment}. Stochastic depth \cite{huang2016deep} and label smoothing \cite{szegedy2016rethinking} are used to regularize. We employ AdamW \cite{loshchilov2017decoupled} optimizer for 300 epochs with a batch size of 1024. The base learning rate of $\frac{\rm {batch\, size}}{512} \times 5 \times 10^{-4}$, 20 epochs of linear warm-up, cosine decay for learning rate, and weight decay of 0.05 are used. The implementation is based on \texttt{PyTorch} \cite{paszke2019pytorch} and \texttt{timm} \cite{rw2019timm}. The details of the hyperparameters are presented in the appendix.
We compare the proposed models with various family models, including CNN-based like ConvNeXt~\cite{liu2022convnet} and ConvFormer~\cite{yu2022metaformer}, attention-based like DeiT~\cite{touvron2020training}, CSwin~\cite{dong2022cswin}, MViTv2~\cite{li2022mvitv2}, DiNAT~\cite{hassani2022dilated}, DaViT~\cite{ding2022davit}, GCViT~\cite{hatamizadeh2023global}, and MaxViT~\cite{tu2022maxvit}, Retention-based like RMT~\cite{fan2023rmt}, MLP-based like DynaMixer~\cite{wang2022dynamixer} and AMixer~\cite{rao2022amixer}, FFT-based like GFNet~\cite{rao2021global}, and hybrid models like CAFormer~\cite{yu2022metaformer}. Table~\ref{tab:imagenet} shows the results. We can see that DFFormers and CDFFormers perform top-1 accuracy among models except for models using attention or retention, with comparable parameters. DFFormers are particularly state-of-the-art among traditional FFT-based models. They perform more than 0.5\% better than other FFT-based models and have been ahead of MLP-based models to which FFT-based tend to compare. CDFFormers also have better cost performance than DFFormers due to using conjunction with convolution. The largest model, CDFFormer-B36, outperforms DFFormer-B36. The above performance comparison of DFFormer and CDFFormer indicates that dynamic filters are promising modules for image recognition.
\subsection{Semantic Segmentation on ADE20K}\label{ssec:ade20k}
\begin{table}[htb]
    \centering
    \small
    \begin{tabular}{lcc}
\toprule
    \multirow{2}{*}{\makecell[c]{Backbone}} & \multirow{2}{*}{\makecell[c]{Prm. \\ (M)}} & \multirow{2}{*}{\makecell[c]{mIoU \\ (\%)}} \\
    ~ & ~ & ~ \\
    \midrule
     ResNet-50~\cite{he2016deep}                & 28.5 &  36.7 \\
     PVT-Small~\cite{wang2021pyramid}          & 28.2 &  39.8 \\
     PoolFormer-S24~\cite{yu2021metaformer}                    & 23.2 &  40.3 \\
     DFFormer-S18 \textbf{(ours)}                   & 31.7 & 45.1  \\
     CDFFormer-S18 \textbf{(ours)}                   & 31.4 & 44.9  \\
    \hline
    ResNet-101~\cite{he2016deep}               & 47.5 &  38.8\\
    ResNeXt-101-32x4d~\cite{xie2017aggregated} & 47.1 &  39.7 \\
    PVT-Medium~\cite{wang2021pyramid}         & 48.0 & 41.6 \\
    PoolFormer-S36~\cite{yu2021metaformer}                    & 34.6 & 42.0 \\
    DFFormer-S36 \textbf{(ours)}                   & 47.2 & 47.5  \\
    CDFFormer-S36 \textbf{(ours)}                   & 46.5 & 46.7  \\
    \hline
    PVT-Large~\cite{wang2021pyramid}          & 65.1 &  42.1 \\
    PoolFormer-M36~\cite{yu2021metaformer}                    & 59.8 & 42.4 \\
    DFFormer-M36 \textbf{(ours)}                   & 66.4 & 47.6  \\
    CDFFormer-M36 \textbf{(ours)}                   & 65.2 & 48.6  \\
\bottomrule
\end{tabular}
    \caption{\label{tab:ade20k}
    \textbf{Performance comparison of models using Semantic FPN trained on ADE20K.} The settings follow \cite{yu2021metaformer} and the results of compared models are cited from \cite{yu2021metaformer}.}
\end{table}
We train and test our models on ADE20K \cite{zhou2017scene} dataset for a semantic segmentation task to evaluate the performance in dense prediction tasks. We employ Semantic FPN \cite{kirillov2019panoptic} in \texttt{mmseg} \cite{mmseg2020} as a base framework. See the appendix for setup details.
Table~\ref{tab:ade20k} shows that DFFormer-based and CDFFormer-based models equipped with Semantic FPN for semantic segmentation are effective for the semantic segmentation task. They are superior to those based on other models, including PoolFormer-based models. From these results, it can be seen that DFFormer-S36 has 5.5 points higher mIoU than PoolFormer-S36. In addition, CDFFormer-M36 achieves 48.6 mIoU. As a result, we also verify the effectiveness of DFFormer and CDFFormer for semantic segmentation.
\subsection{Ablation Studies}\label{ssec:ablation}
\begin{table*}[htb]
    \centering
    \small
    \begin{tabular}{llrrrr}
\toprule
Ablation & Variant & Parameters (M) & FLOPs (G) & Throughput (img/sec) & Top-1 Acc. (\%) \\
\midrule
Baseline & DFFormer-S18 & 30 & 3.8 & 535 & 83.2  \\
\hline
 \multirow{4}{*}{Filter} & $N=4\,\rightarrow\,2$& 29 & 3.8 & 534 & 83.0 \\
 & $\rho=0.25\,\rightarrow\,0.125$& 28 & 3.8 & 532 & 83.1 \\
 & GFFormer-S18 & 30 & 3.8 & 575 & 82.9 \\
 & DF $\rightarrow$ AFNO & 30 & 3.8 & 389 & 82.6 \\
 \hline
\multirow{3}{*}{Activation}  & StarReLU $\rightarrow$ GELU  & 30 & 3.8 & 672 & 82.7 \\
             & StarReLU $\rightarrow$ ReLU & 30 & 3.8 & 640 & 82.5 \\
             & StarReLU, DF$\rightarrow$ ReLU, AFNO & 30 & 3.8 & 444 & 82.0 \\

\bottomrule
\end{tabular}
    \caption{\label{tab:ablation_former}
    Ablation for DFFormer-S18 on ImageNet-1K. Remind $N$ is the dimension of the dynamic filter basis, and $\rho$ is a ratio of MLP intermediate dimension to input dimension.}
\end{table*}
\begin{figure}[htb]
  \centering
  \includegraphics[width=1.\linewidth]{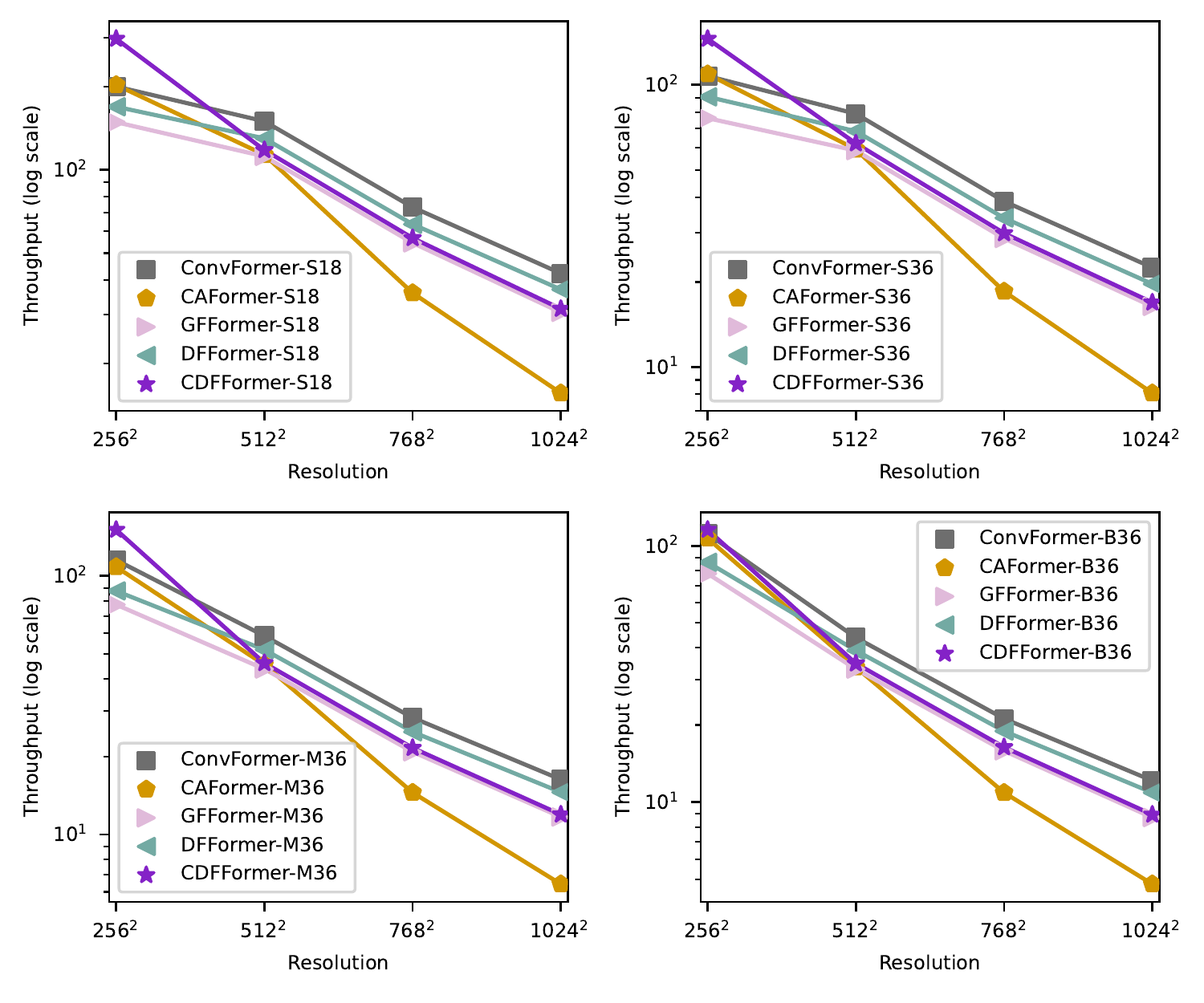}
\caption{\label{fig:throughput}\textbf{Throughput vs. resolution.} Throughput has been benchmarked on a V100 with 16GB memory at a batch size of four.}
\end{figure}
\begin{figure}[htb]
  \centering
  \includegraphics[width=1.\linewidth]{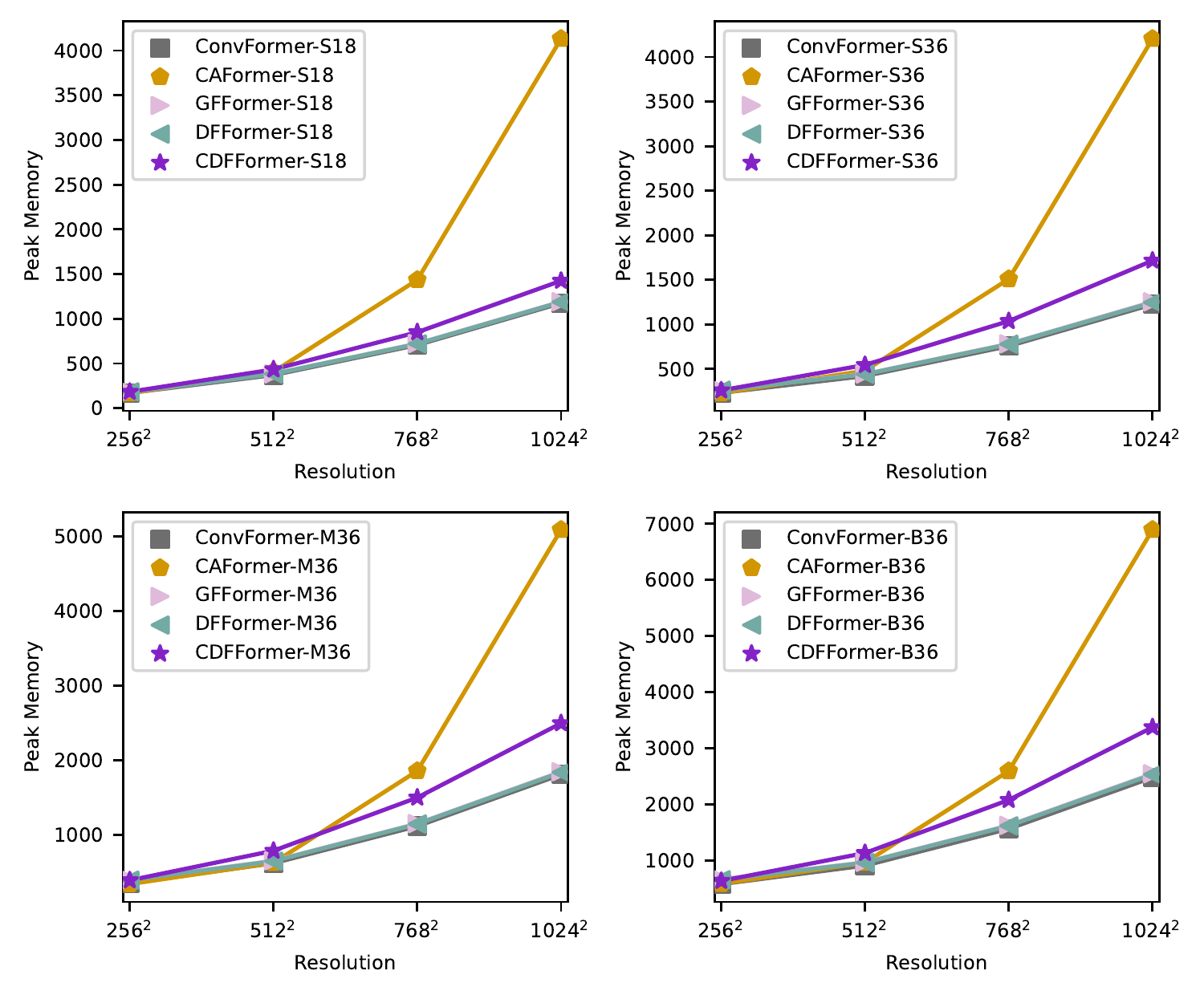}
\caption{\label{fig:memory}\textbf{Peak memory vs. resolution.} Peak memory has been benchmarked on a V100 with 16GB memory at a batch size of four.}
\end{figure}
We experiment with ablation studies on the ImageNet-1K dataset. Let us discuss ablation from several perspectives.
\paragraph{Filter} We studied how changing filters would work. First, we change to the hyperparameters of dynamic filters. Specifically, we train and test models in which the dimension of dynamic filter base $N$ and the intermediate dimension $\rho C$ are modified in half of the cases, respectively. From Table~\ref{tab:ablation_former}, however, we found that the throughput and number of parameters remained almost the same, and the accuracy dropped by 0.2\% and 0.1\%, respectively. Next, we train GFFormer-S18, which is replaced by each dynamic filter of DFformer-S18 with a global filter (i.e., static filter) to confirm the effectiveness of dynamic filters. Table~\ref{tab:ablation_former} demonstrates DFformer-S18 is superior to GFFormer-S18 0.3\%. We also have studied a case where AFNOs replaced dynamic filters but found that the AFNO-based model degraded in accuracy more than the dynamic filter and that dynamic filters have better throughput than AFNOs.
\paragraph{Activation} DFFormer and CDFFormer use StarReLU, but this is a relatively new activation; many MetaFormers employ Gaussian Error Linear Unit (GELU) \cite{hendrycks2023gaussian}, and earlier generations of convolutional neural networks (CNNs), such as ResNet \cite{he2016deep}, often used Rectified Linear Unit (ReLU) \cite{nair2010rectified}. Accordingly, we also experiment with a version of the model in which GELU and ReLU replace StarReLU in DFFormer-S18. As a result, Table~\ref{tab:ablation_former} shows that DFFormer-S18 outperforms the replaced version of the model by more than 0.5\%, demonstrating that StarReLU is also a valuable activation for dynamic filters. In this manner, StarReLU has a significant impact on the final performance. Therefore, to comprehend how much a dynamic filter would improve performance without StarReLU, we also experimented with a model version in which StartReLU and a dynamic filter are replaced by ReLU and AFNO, respectively. Comparing this result with the StarReLU $\to$ ReLU results shows us that the dynamic filter has enough impact even when the influence of StarReLU is removed.

\section{Analysis}
\subsection{Advantages at Higher Resolutions}\label{ssec:advantages}
We observe how the throughput and peak memory at inference change on proposed and comparative models when resolution varies. For throughput, the proposed models have been inferior to CAFormer, an architecture that utilized MHSA, at a resolution of $224^2$ since Table~\ref{tab:imagenet} can comprehend the fact. The result conflicts with the computational complexity presented in the Introduction. Our interpretation of the issue is that the actual throughput would depend on the implementation of FFT (although we use \texttt{cuFFT} via \texttt{PyTorch}), hardware design, etc. Increasing the resolution, the theoretically computational complexity of MHSA comes into effect: Figure~\ref{fig:throughput} shows the throughput for different resolutions. DFFormer and CDFFormer maintain throughput close to that of ConvFormer, while only CAFormer shows a significant decrease in throughput. The same is true for peak memory in Figure~\ref{fig:memory}: Whereas the CAFormer peak memory increases, the DFFormer and CDFFormer peak memory is comparable to that of ConvFormer. Therefore, DFFormer and CDFFormer are beneficial in speed- and memory-constrained environments for tasks such as semantic segmentation, where high resolution is required.
\subsection{Representational Similarities}
\begin{figure}[!thb]
\centering
\begin{minipage}[b]{0.3\hsize}
  \centering
  \includegraphics[width=1.\linewidth, trim=70 30 10 30]{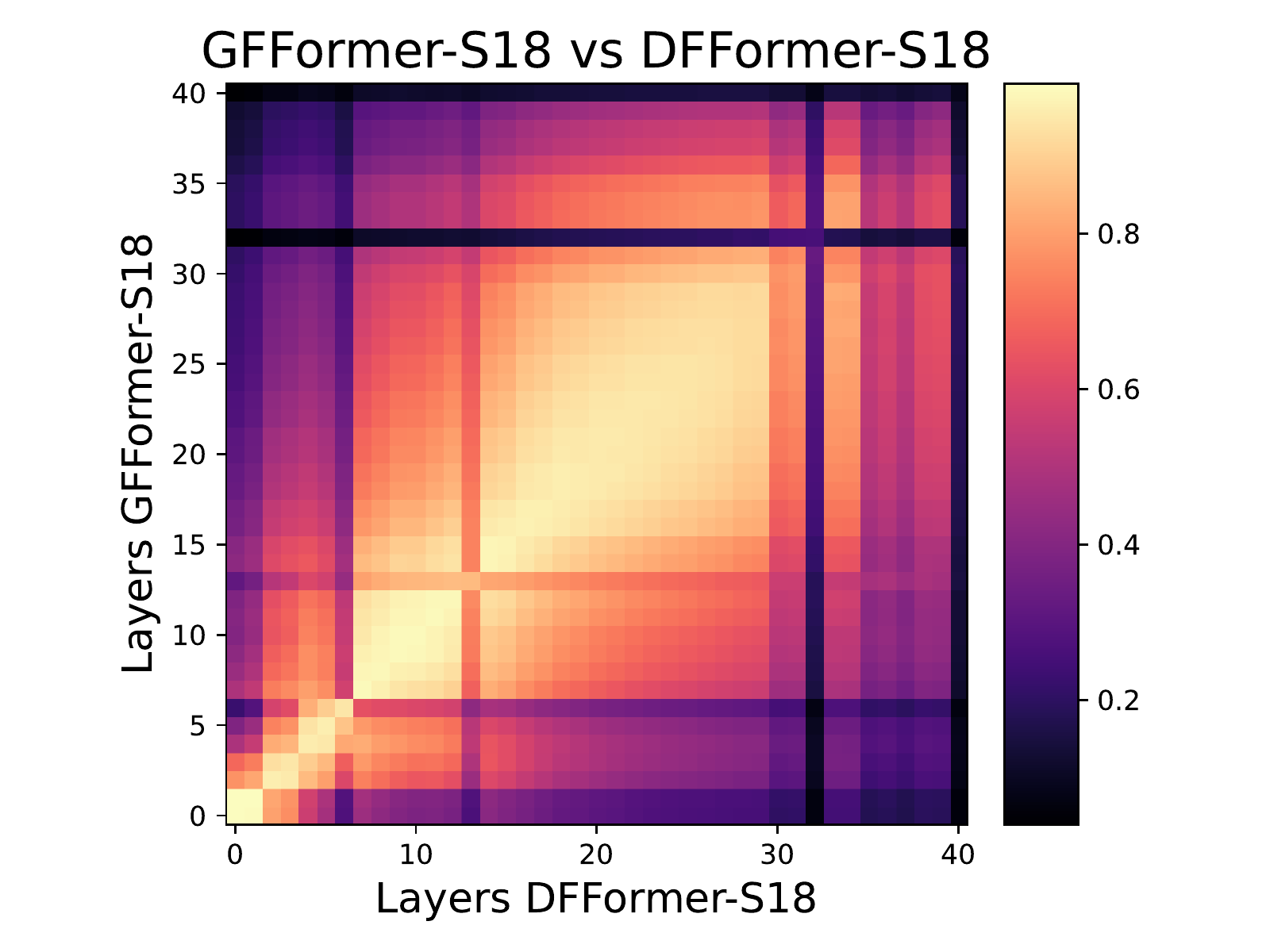}
\subcaption{\label{fig:gfformer_s18_dfformer_s18_cka}GF vs. DF }
\end{minipage}
\begin{minipage}[b]{0.3\hsize}
  \centering
  \includegraphics[width=1.\linewidth, trim=70 30 10 30]{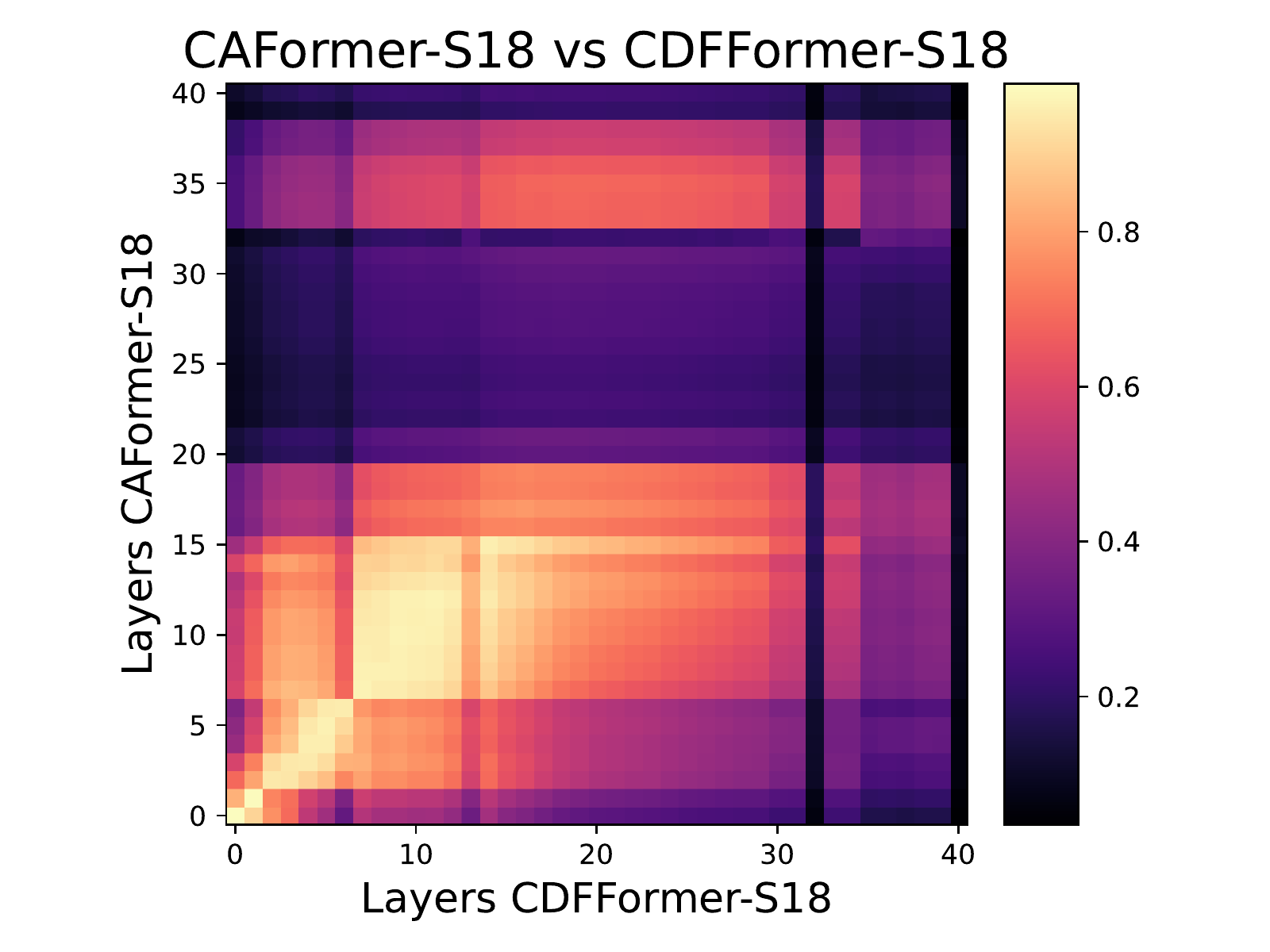}
\subcaption{\label{fig:caformer_s18_cdfformer_s18_cka}CA vs. CDF}
\end{minipage}
\begin{minipage}[b]{0.3\hsize}
  \centering
  \includegraphics[width=1.\linewidth, trim=70 30 10 30]{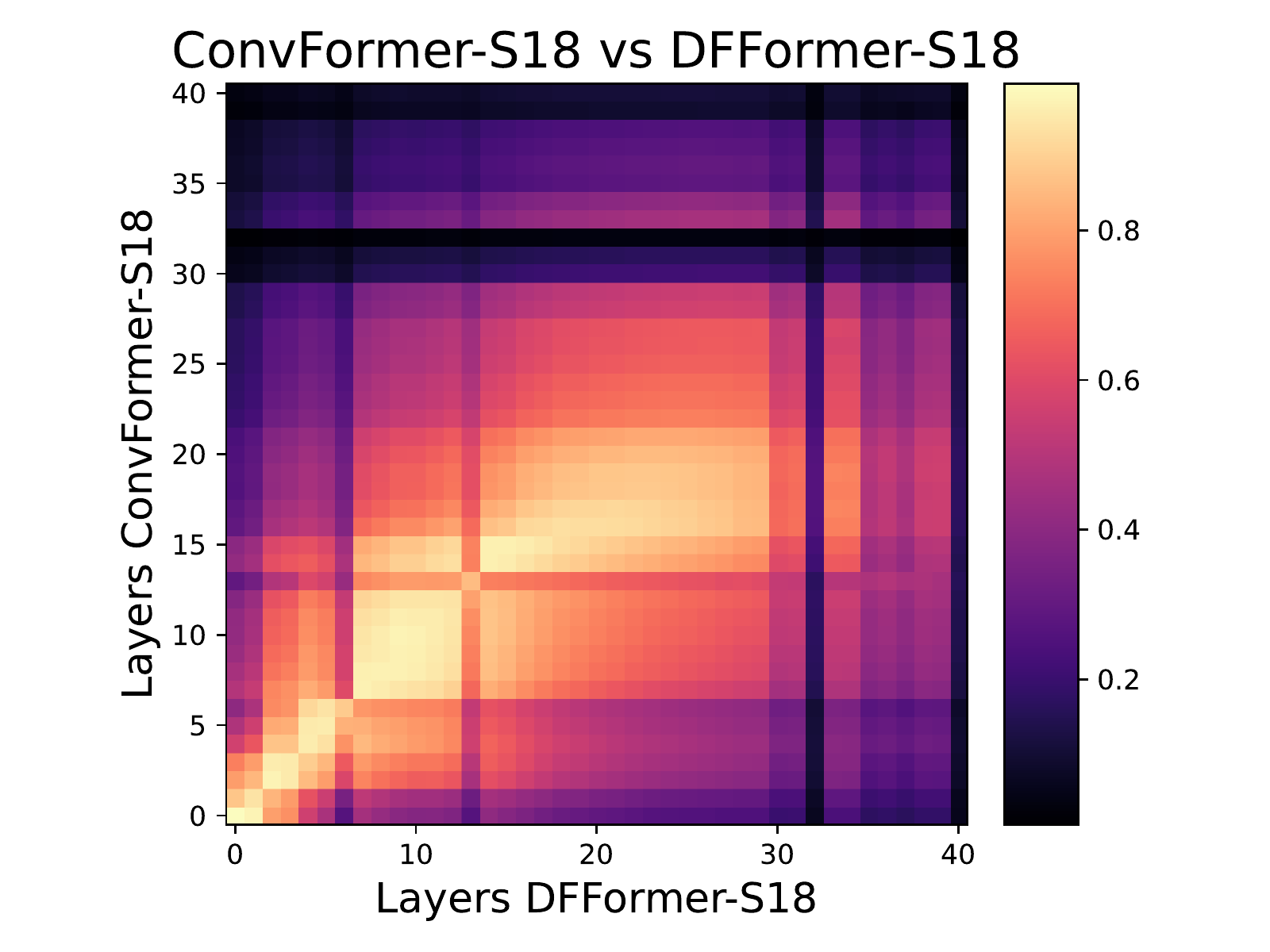}
\subcaption{\label{fig:convformer_s18_dfformer_s18_cka}Conv vs. DF}
\end{minipage}
\caption{\label{fig:cka}The feature map similarities by CKA}
\end{figure}

Figure~\ref{fig:cka} shows the similarities between GFFormer-S18 and DFFormer-S18, ConvFormer-S18 and DFFormer-S18, and CAFormer-S18 and CDFFormer-S18 in the validation set of ImageNet-1K. In our analysis, we metric the similarity using mini-batch linear CKA \cite{nguyen2020wide}. \texttt{torch\_cka} \cite{subramanian2021torch_cka} toolbox was used to implement the mini-batch linear CKA. The layers to be analyzed included four downsampled and 18 token-mixers residual blocks and 18 channel MLP residual blocks, and the indices in the figure are ordered from the shallowest layer. Figure~\ref{fig:gfformer_s18_dfformer_s18_cka} shows that GFFormer-S18 and DFFormer-S18 are very similar up to stage 3, although they are slightly different at stage 4. We also found the same thing about the similarity between ConvFormer-S18 and DFFormer-S18 from Figure~\ref{fig:convformer_s18_dfformer_s18_cka}. On the contrary, Figure~\ref{fig:caformer_s18_cdfformer_s18_cka} shows that CAFormer-S18 and CDFFormer-S18 are similar up to stage 2 with convolution but become almost entirely different from stage 3 onward. Even though FFT-based token-mixers, like MHSA, are token-mixers in the global domain and can capture low-frequency features, they necessarily do not learn similar representations. The fact would hint that MHSA and the FFT-based token-mixer have differences other than spatial mixing.
\begin{figure}[thb]
\centering
\begin{minipage}[t]{0.48\hsize}
  \centering
  \includegraphics[width=1.\linewidth]{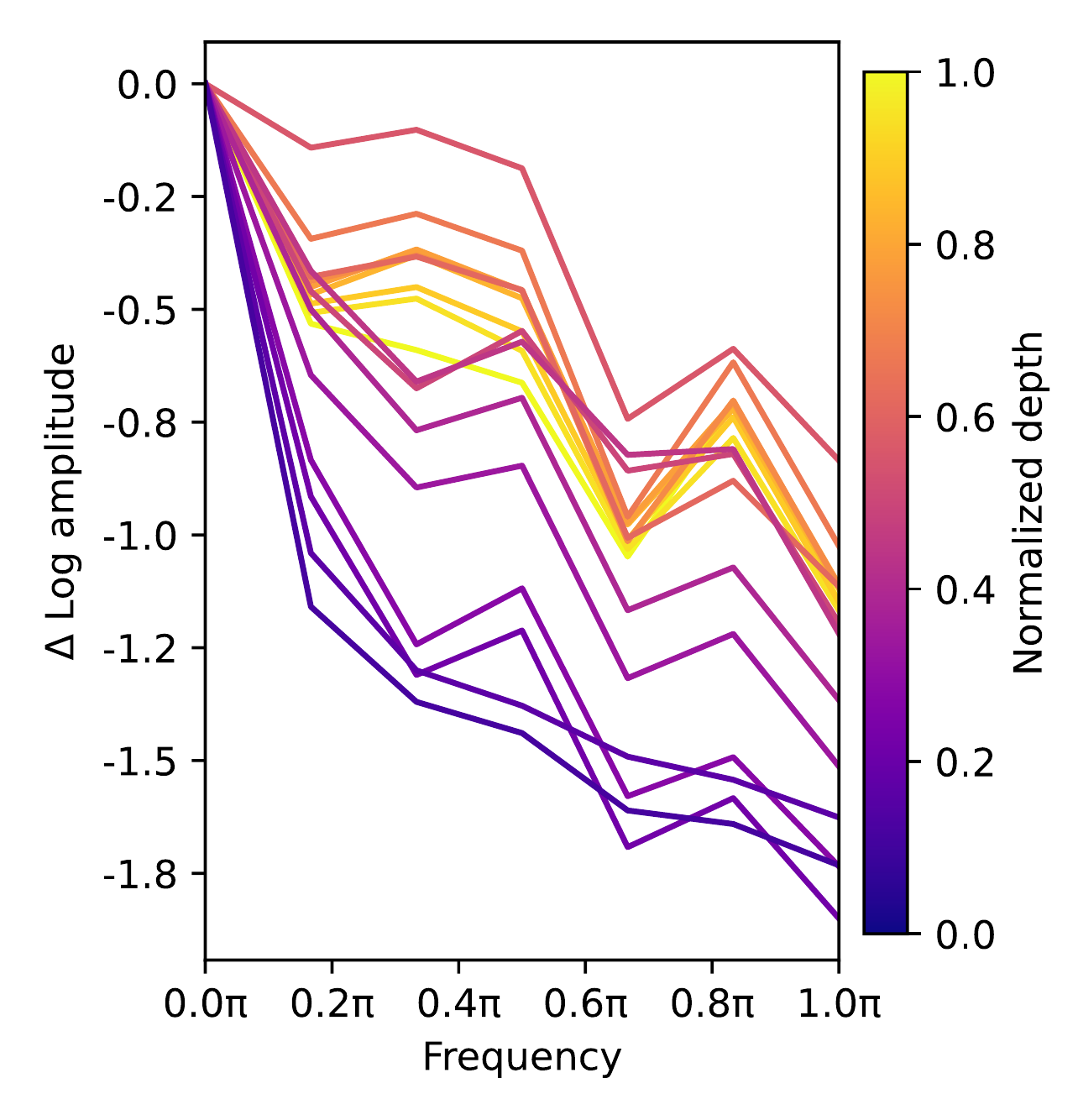}\subcaption{\label{fig:logamp_caformer}CAFormer-S18}
\end{minipage}
\begin{minipage}[t]{0.48\hsize}
  \centering
  \includegraphics[width=1.\linewidth]{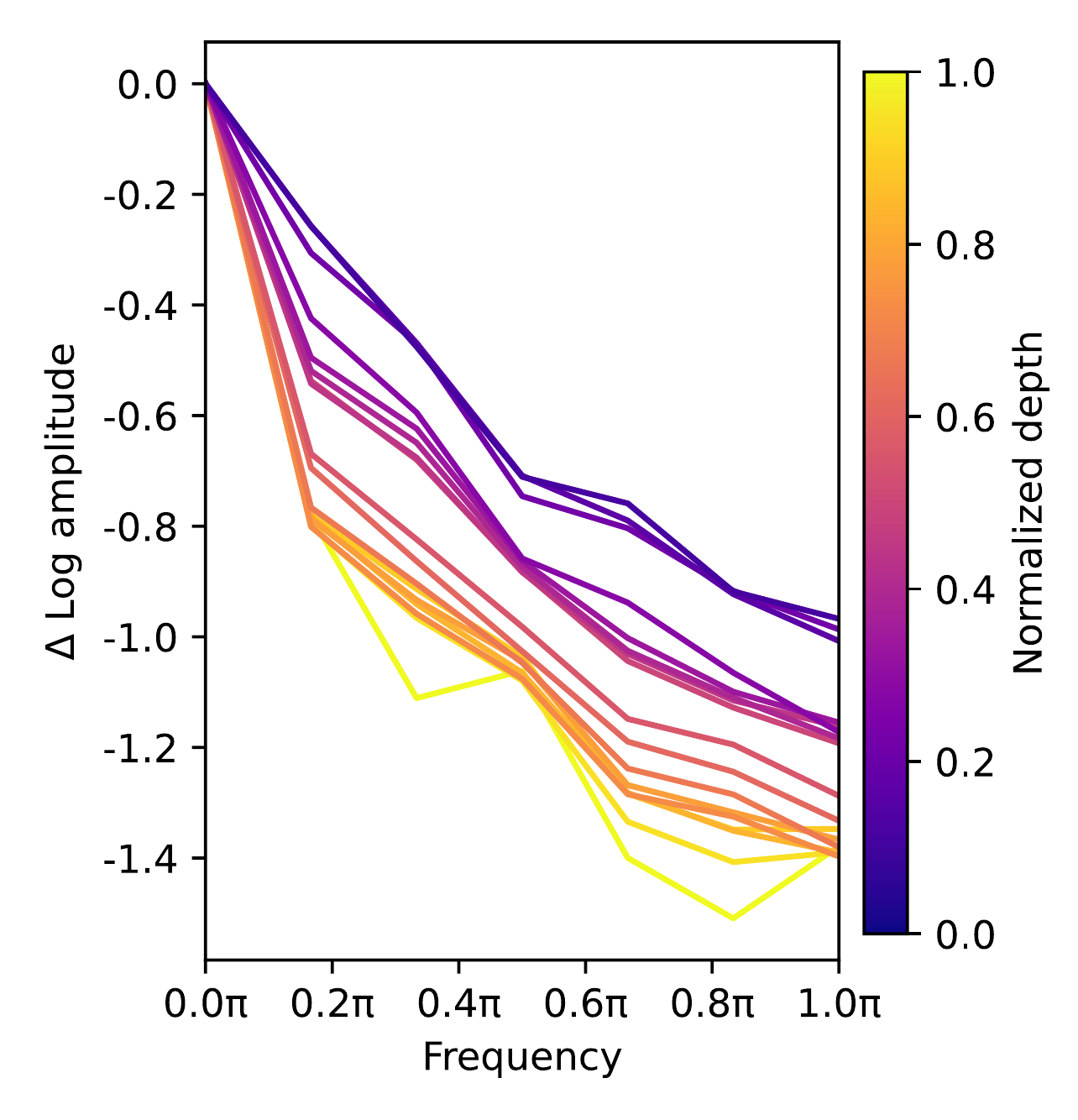}
  \subcaption{\label{fig:logamp_cdfformer}CDFFormer-S18}
\end{minipage}
\caption{\label{fig:logamp_former}Relative log amplitudes of Fourier transformed feature maps on stage 3. A setup similar to that in Table~\ref{fig:logamp}.}
\end{figure}
We perform a Fourier analysis on stage 3 to know the difference in stage 3. Figure~\ref{fig:logamp_former} shows the difference between MHSA and FFT-based token-mixer results. Surprisingly, CAFormer-S18 acts as a high-pass filter, whereas the result for CDFFormer-S18 indicates more attenuated at other frequencies than around 0. In other words, MHSA incorporated in a hierarchical architecture can learn high-pass and low-pass filters while the dynamic filter attenuates the high frequencies. Only applying to stage 3 makes the analysis more interpretable than overall because this analysis drastically changes frequency by downsampling.
\subsection{Analysis of Dynamic Filter Basis}\label{ssec:visualization}
\begin{figure}[thb]
\begin{minipage}[b]{1.\hsize}
  \centering
  \includegraphics[width=1.\linewidth]{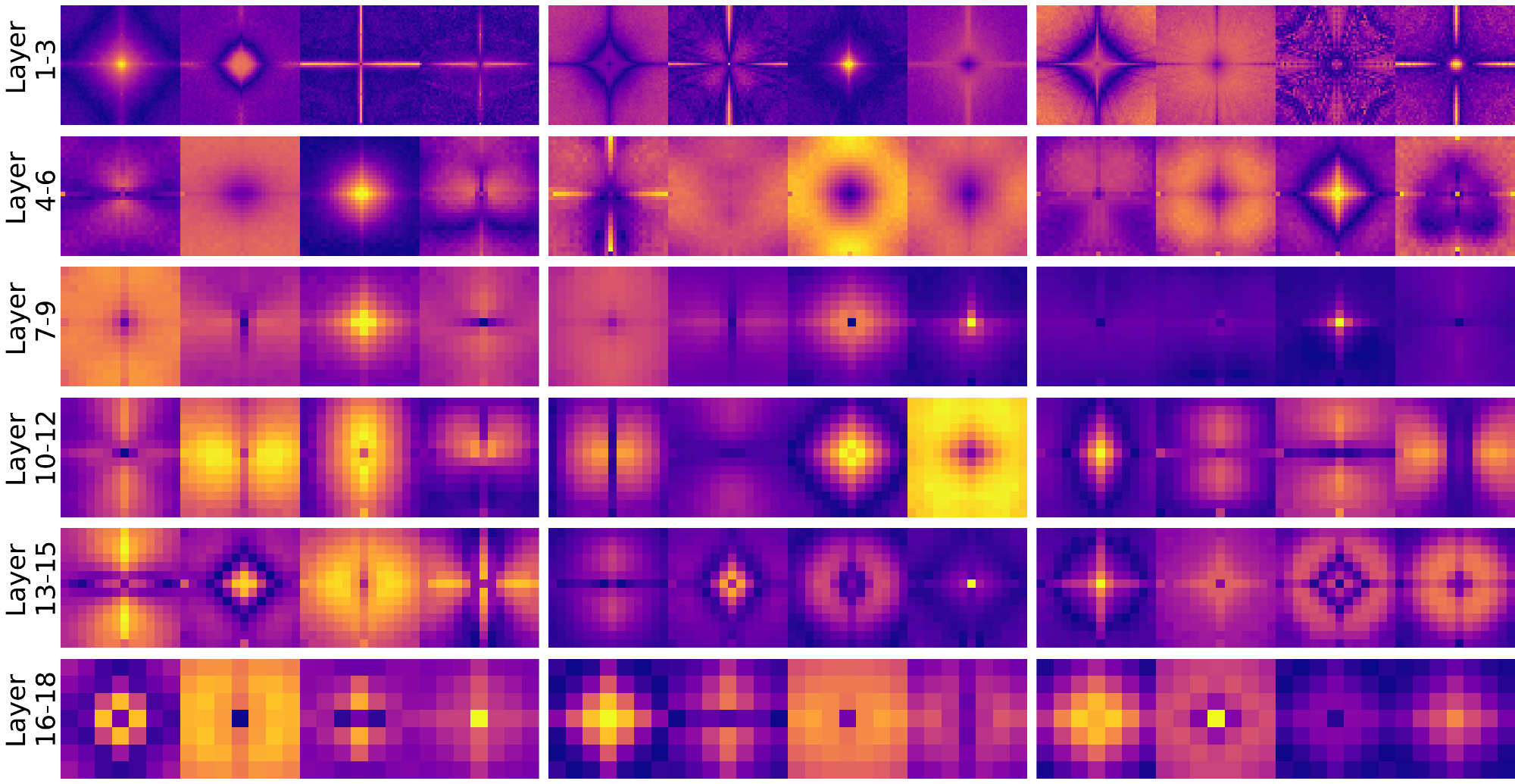}
  \caption{\label{fig:dfformers18_filter}Filters in the frequency domain on DFFormer-S18}
\end{minipage}
\end{figure}

The basis of complex parameters in the frequency domain can represent dynamic filters used in DFFormer and CDFFormer. Here, we visualize them for DFFormer-S18, which has four in each dynamic filter module. In our visualization, the center pixel means zero frequency and the yellower pixel means higher amplitudes. Figure~\ref{fig:dfformers18_filter} demonstrates that the redundancy of filters in the same layer is reduced compared to Figure~\ref{fig:gfnet_ti_filter} about GFNet-Ti. In addition, we can see high-pass, low-pass, and band-pass filters in many layers. Details of the visualization method and more visualization will be given in the appendix.
\label{ssec:shape_bias}
\section{Conclusion}
This paper studied the similarities and differences between the global filter and MHSA in vision models. According to the analysis, we proposed a novel dynamic filter responsible for dynamically generating global filters. Based on this module, we have developed new MetaFormer classes, DFFormer and CDFFormer, which achieve promising performance in MHSA-free MetaFormers. We have proven through a variety of experiments that the proposed models perform impressively not only in image classification but also in downstream tasks. In addition, our models can process high-resolution images faster and with less memory than MetaFormers using MHSA. We also found that the representation and properties of our models are not similar unexpectedly to that of models using MHSA.

\section{Acknowledgments}
This work was supported by the Rikkyo University Special Fund for Research.

\bibliography{aaai24}

\appendix
\section{More Experiment}
\subsection{Object Detection on COCO}\label{ssec:coco}
\begin{table*}[htb]
    \centering
    \begin{NiceTabular}{l|ccccccc}
\toprule
Backbone & Parameters(M) & AP(\%) & AP$_{50}$ & AP$_{75}$(\%) & AP$_S$(\%) & AP$_M$(\%) & AP$_L$(\%) \\
\midrule
ResNet-50  & 37.7 & 36.3 & 55.3 & 38.6 & 19.3 & 40.0 & 48.8 \\
PoolFormer-S24          & 31.1 & 38.9 & 59.7 & 41.3 & 23.3 & 42.1 & 51.8  \\
DFFormer-S18 \textbf{(ours)}          & 38.1 & 43.6 & 64.5 & 46.6 & 27.5 & 47.3 & 58.1  \\ 
CDFFormer-S18 \textbf{(ours)}          & 37.4 & 43.4 & 64.7 & 46.3 & 26.3 & 47.1 & 57.3  \\ 
\hline
ResNet-101  & 56.7 & 38.5 & 57.8 & 41.2 & 21.4 & 42.6 & 51.1 \\
PoolFormer-S36           & 40.6 & 39.5 & 60.5 & 41.8 & 22.5 & 42.9 & 52.4 \\ 
DFFormer-S36 \textbf{(ours)}          & 53.8 & 45.3 & 66.1 & 48.7 & 26.9 & 49.0 & 59.9 \\ 
CDFFormer-S36 \textbf{(ours)}          & 52.6 & 45.0 & 66.0 & 47.8 & 27.6 & 48.5 & 59.6 \\ 
\bottomrule
\end{NiceTabular}
    \caption{\label{tab:coco}
    \textbf{Performance comparison of models using RetinaNet trained on COCO.} The settings follow \cite{yu2021metaformer} and the results of compared models are cited from \cite{yu2021metaformer}.}
\end{table*}
We evaluate the performance of our models on downstream tasks, especially in object detection on COCO benchmark \cite{lin2014microsoft}. It has 80 classes consisting of 118,287 training images and 5,000 validation images. We employ RetinaNet as an object detection framework. The implementation takes \cite{lin2017focal} of \texttt{mmdet} \cite{chen2019mmdetection}. We utilize pre-trained models on ImageNet-1K dataset as the backbones. Following the setting in \cite{yu2021metaformer}, we use 1$\times$ training, meaning 12 epochs, batch size of 16, and AdamW \cite{loshchilov2017decoupled} optimizer with an initial learning rate of $10^{-4}$. Similar to the general setup, training images of coco are resized to no more than 800 pixels on the short side and 1,333 pixels on the long side, keeping the aspect ratio. The testing images are also resized to 800 pixels on the short side. The default dynamic filter does not support arbitrary resolutions because of using the element-wise product. Although input images have an indeterminate resolution in the setting, pre-trained DFFormer and CDFFormer are bound with the resolution of $224^2$. To cancel this problem, we bicubic interpolate global filter basis $\mathbb{K}$ to fit images with a resolution of $800^2$ and let them be new parameters. Moreover, the global filter basis is interpolated to adjust to the shape of frequency features input to dynamic filters. 

Table~\ref{tab:coco} shows RetinaNets with DFFormer and CDFFormer as backbones outperform comparable ResNet and PoolFormer backbones. For instance, DFFormer-S36 has 5.8 points higher AP than PoolFormer-S36. Thus, DFFormer and CDFFormer obtained competitive results in COCO object detection.
\subsection{Settings of Semantic Segmentation on ADE20K}\label{ssec:settings_ade20k}
We remark on the datasets and settings of the semantic segmentation task experiments described in the main text. We use ADE20K \cite{zhou2017scene} dataset, which has 150 semantic categories, 20,210 images for training, 2,000 for validation, and 3,000 for testing. We employ Semantic FPN \cite{kirillov2019panoptic} in \texttt{mmseg} \cite{mmseg2020} as a base framework. Training images are resized and cropped to a shape of $512^2$. For testing, images are resized to a shorter size of 512 pixels, keeping the aspect ratio. We adopt global filter basis parameters $\mathbb{K}$ interpolated from pre-trained parameters to suit images with a resolution of $512^2$. For testing, $\mathbb{K}$ is also interpolated for the same reason as subsection~\ref{ssec:coco}. We utilize AdamW \cite{loshchilov2017decoupled} optimizer, initial learning rate of $2\times 10^{-4}$, polynomial scheduler with a power 0.9, batch size of 32, and 40K iterations, following \cite{yu2021metaformer}.
\section{More Analysis}
\subsection{More Visualization of Dynamic Filter Basis}
\begin{figure*}[b]
\begin{minipage}[t]{0.48\hsize}
\begin{minipage}[t]{1.\hsize}
  \centering
  \includegraphics[width=1.\linewidth]{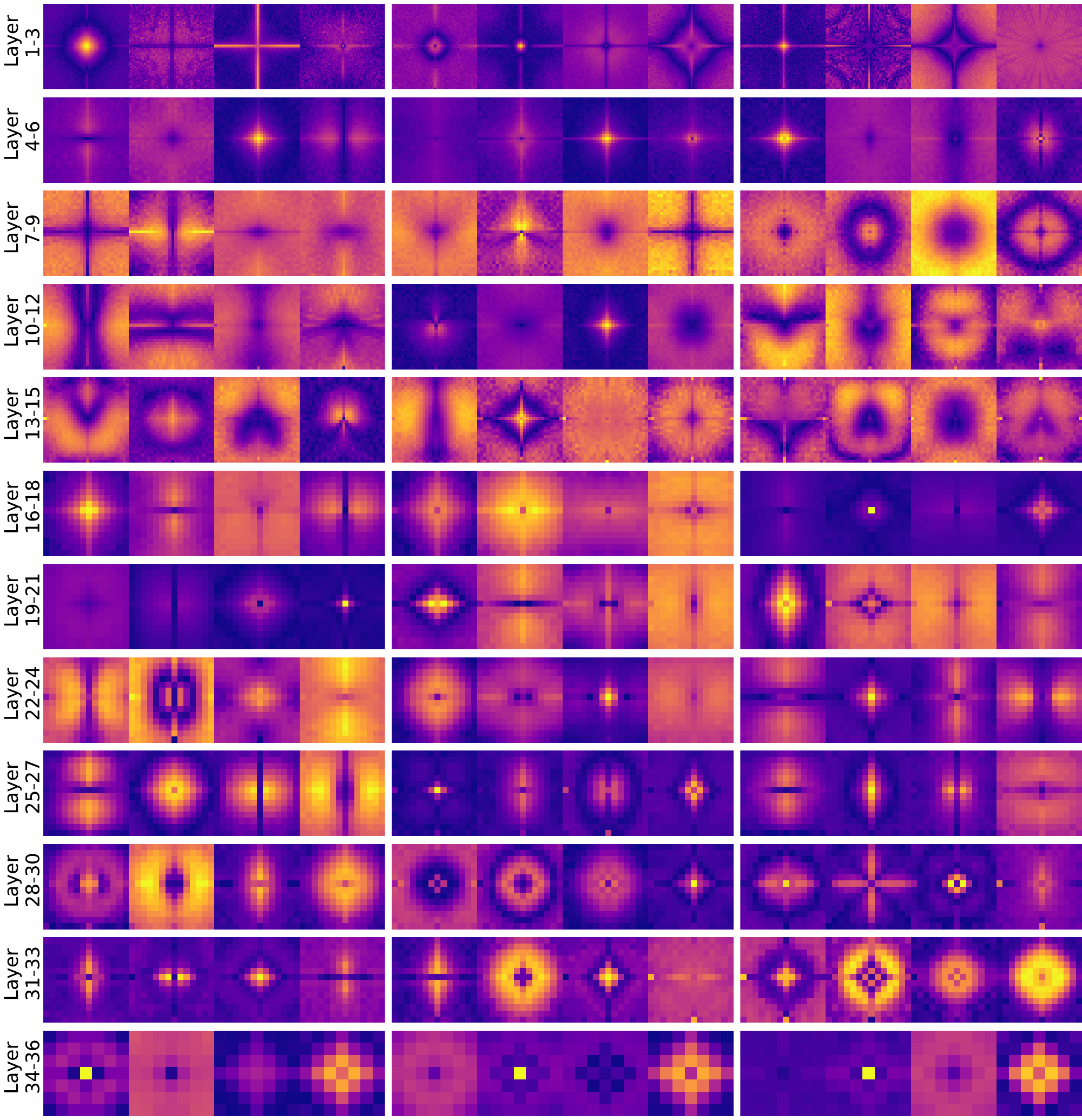}
\subcaption{\label{fig:dfformer_s36_filter}DFFormer-S36}
\end{minipage}
\begin{minipage}[t]{1.\hsize}
  \centering
  \includegraphics[width=1.\linewidth]{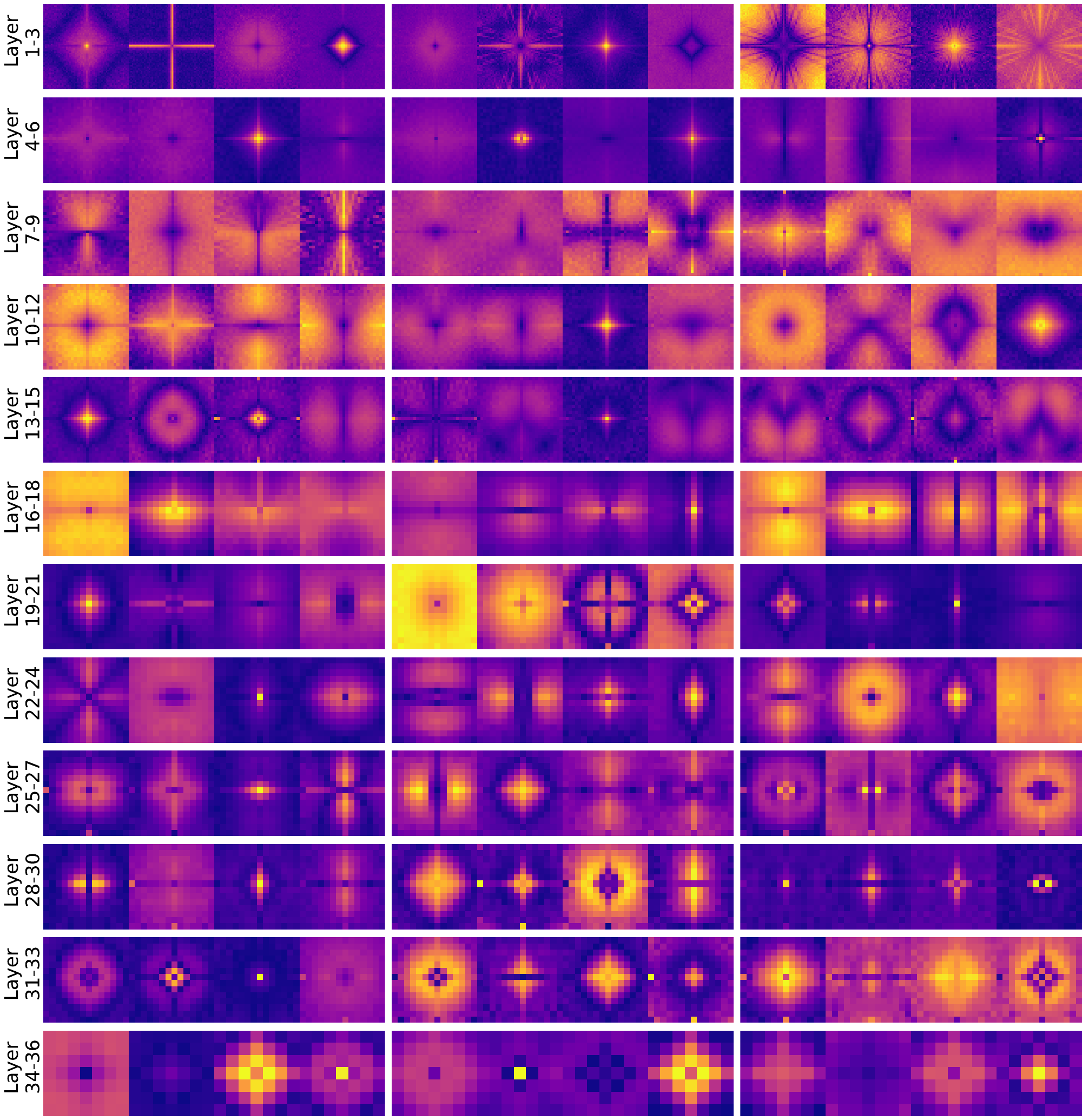}
\subcaption{\label{fig:dfformer_m36_filter}DFFormer-M36}
\end{minipage}
\end{minipage}
\begin{minipage}[t]{0.48\hsize}
\begin{minipage}[t]{1.\hsize}
  \centering
  \includegraphics[width=1.\linewidth]{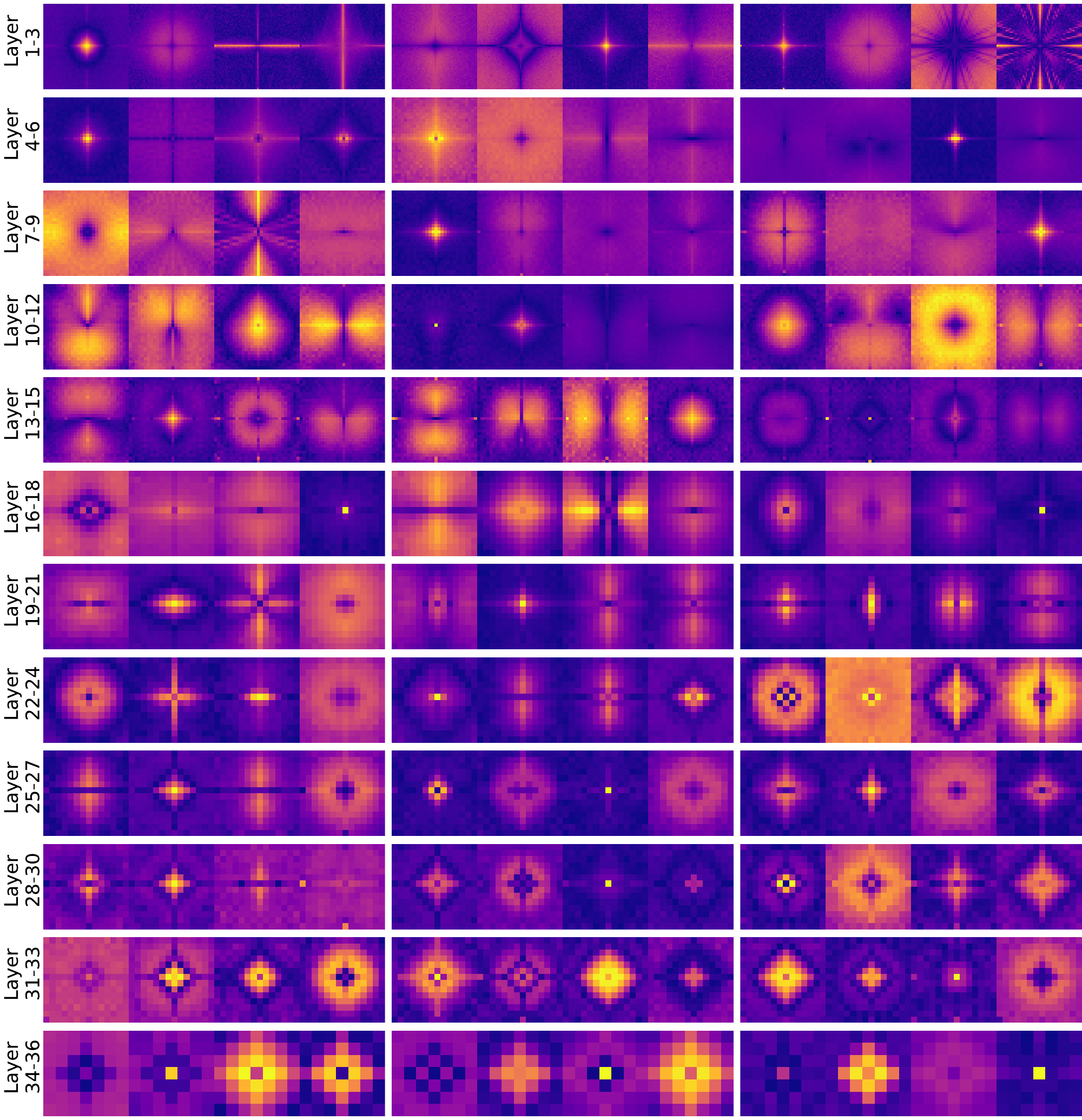}
\subcaption{\label{fig:dfformer_b36_filter}DFFormer-B36}
\end{minipage}
\end{minipage}
\caption{Visualization of dynamic filter basis in the frequency domain on DFFormers}\label{fig:dfformer_filter}
\end{figure*}

\begin{figure*}[b]
\begin{minipage}[t]{0.48\hsize}
\begin{minipage}[t]{1.\hsize}
  \centering
  \includegraphics[width=1.\linewidth]{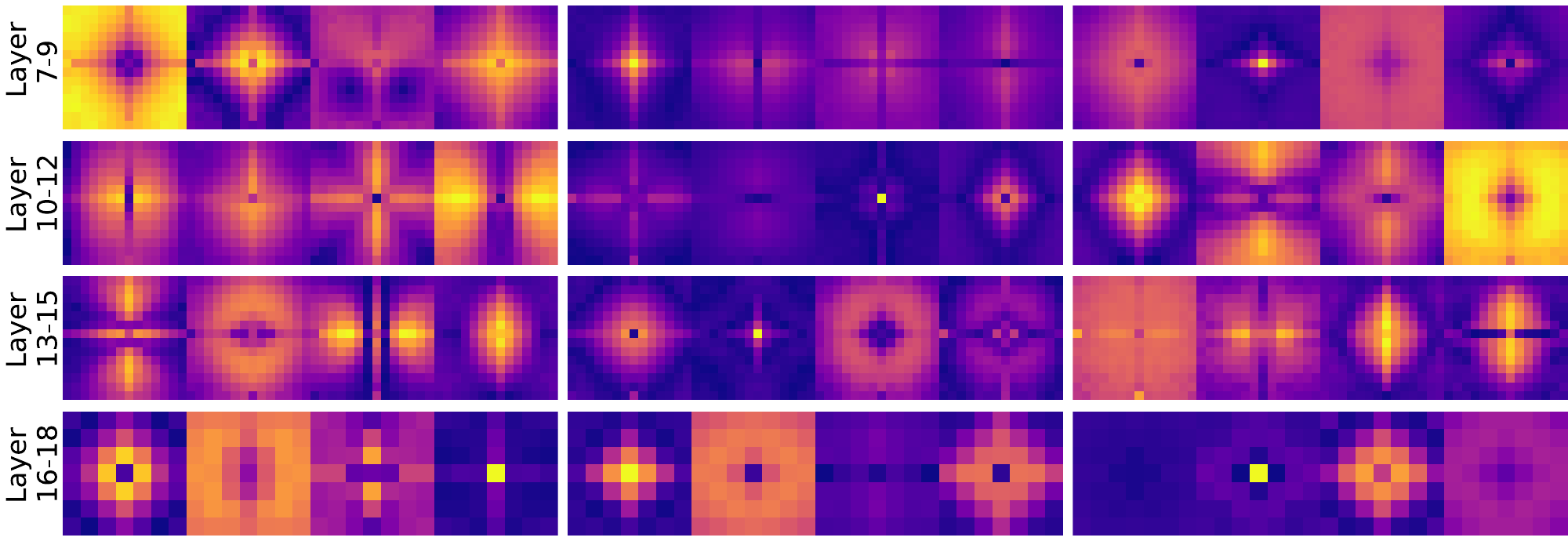}
\subcaption{\label{fig:cdfformer_s18_filter}CDFFormer-S18}
\end{minipage}
\begin{minipage}[t]{1.\hsize}
  \centering
  \includegraphics[width=1.\linewidth]{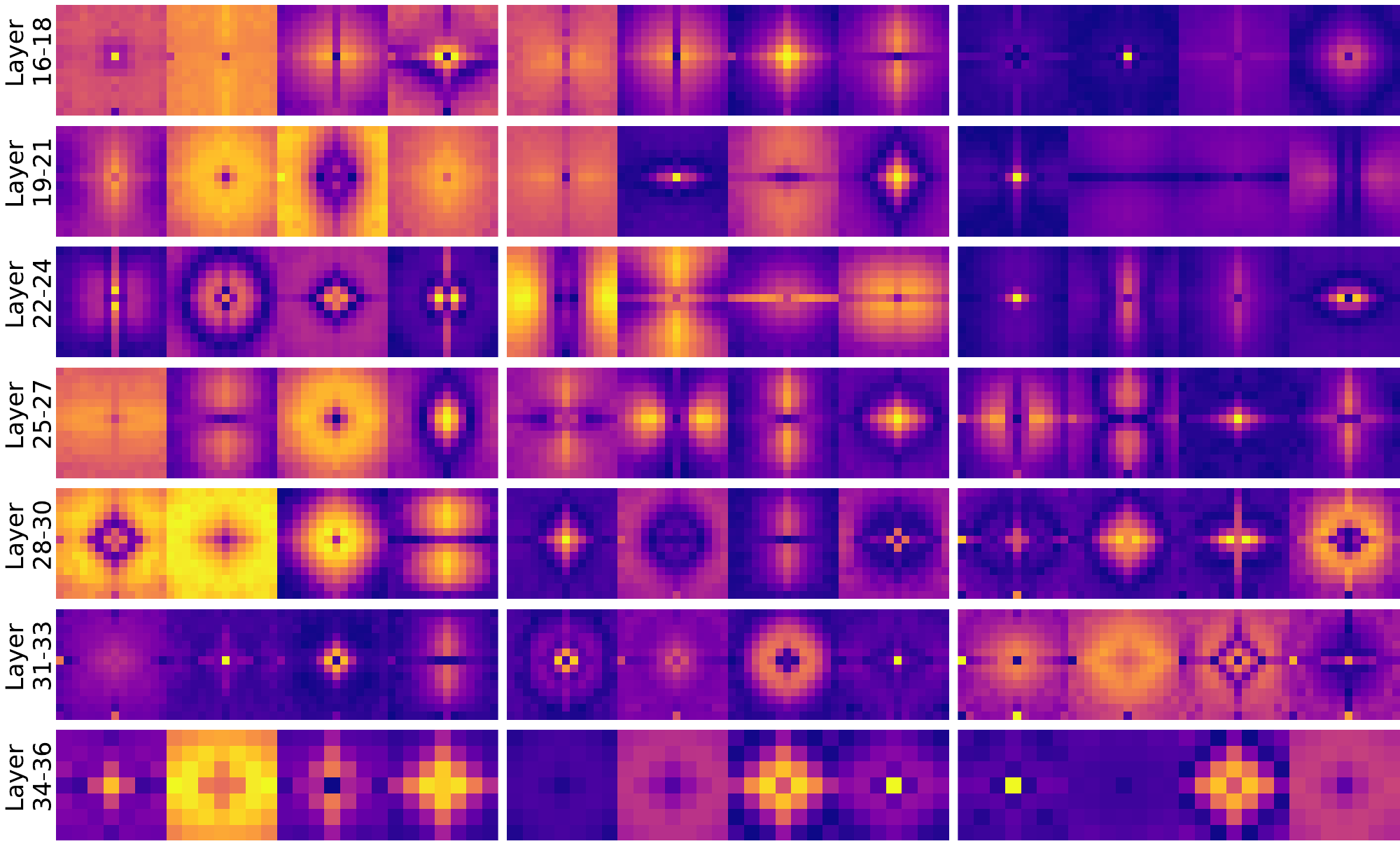}
\subcaption{\label{fig:cdfformer_s36_filter}CDFFormer-S36}
\end{minipage}
\begin{minipage}[t]{1.\hsize}
  \centering
  \includegraphics[width=1.\linewidth]{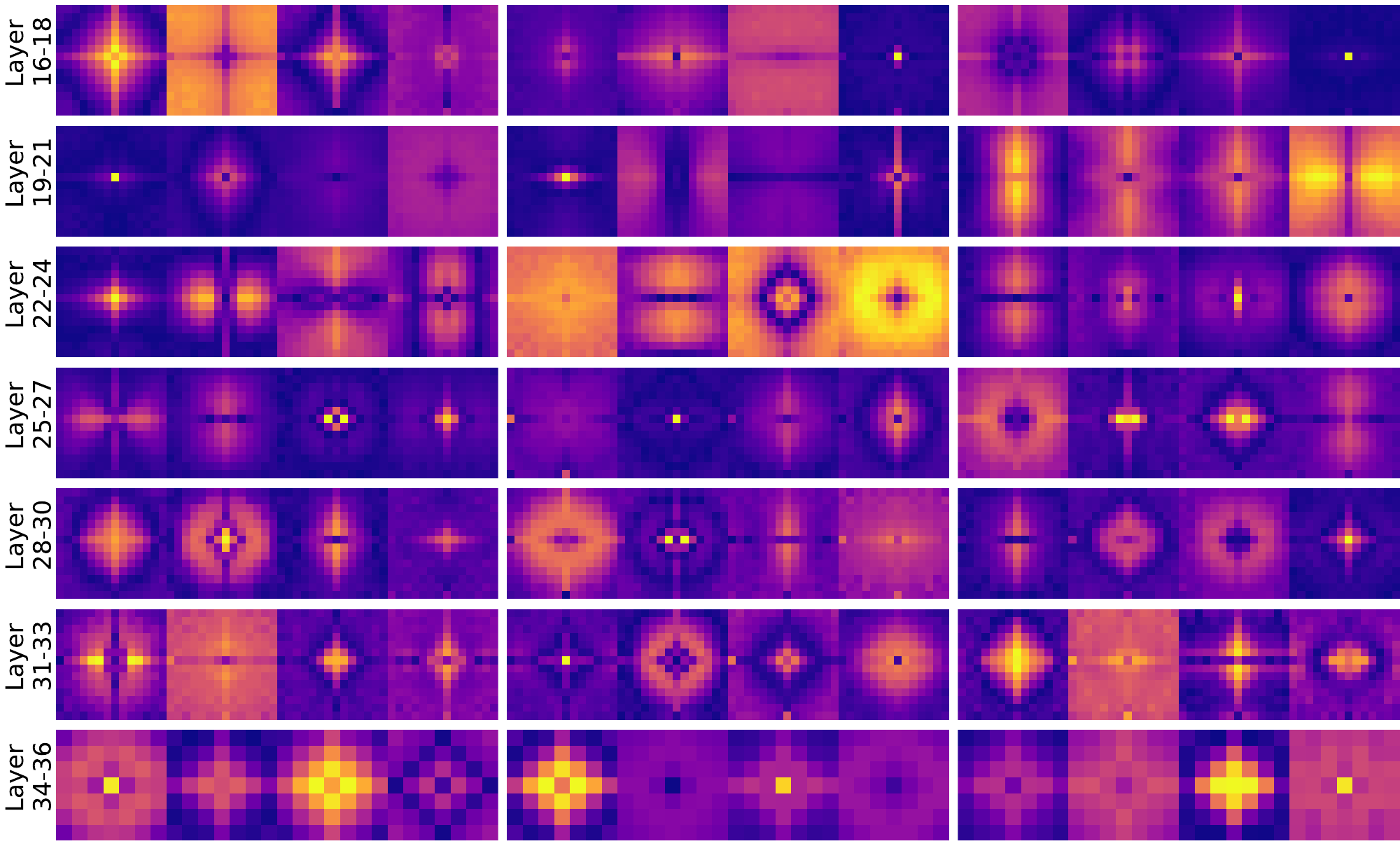}
\subcaption{\label{fig:cdfformer_m36_filter}CDFFormer-M36}
\end{minipage}
\end{minipage}
\begin{minipage}[t]{0.48\hsize}
\begin{minipage}[t]{1.\hsize}
  \centering
  \includegraphics[width=1.\linewidth]{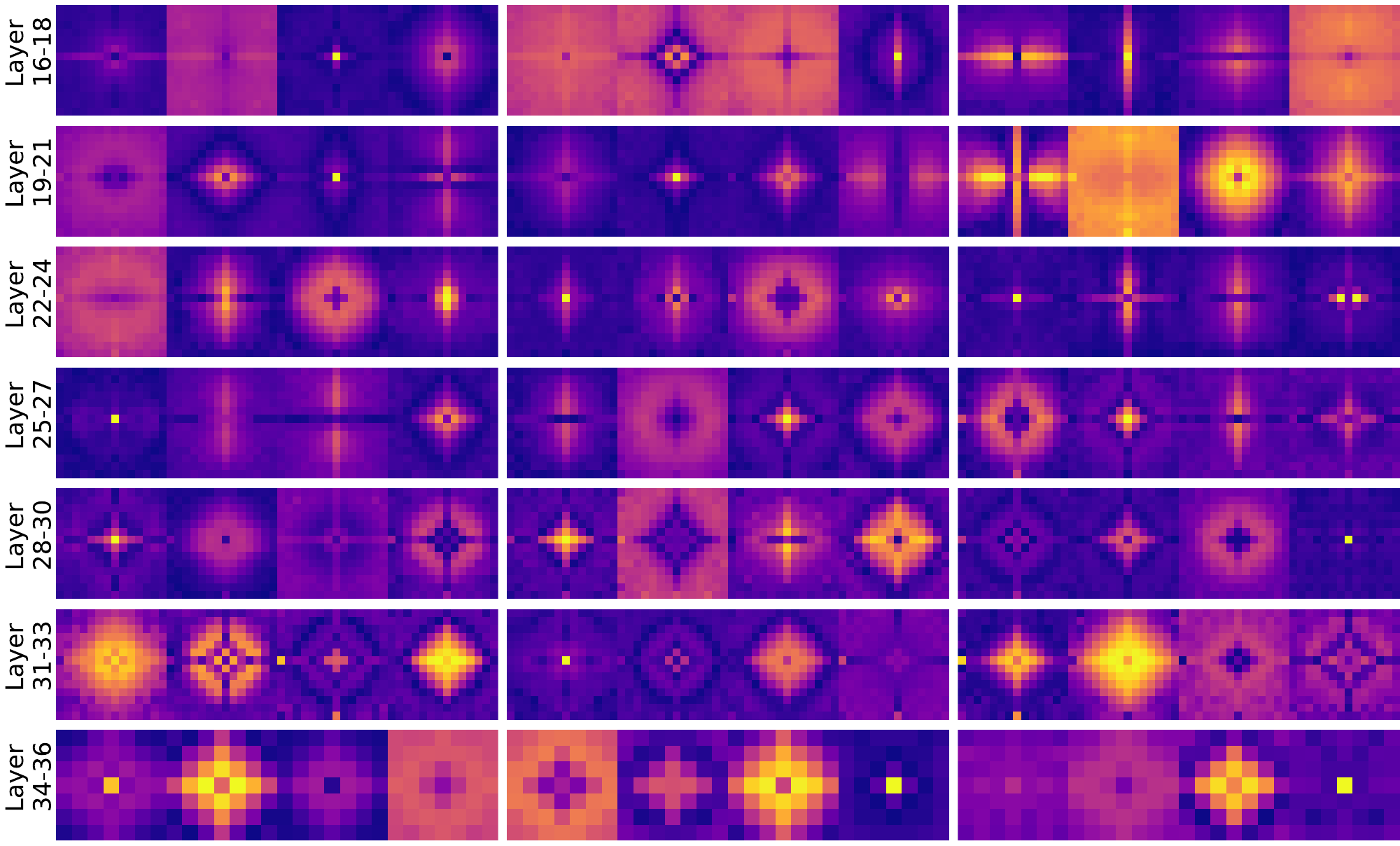}
\subcaption{\label{fig:cdfformer_b36_filter}CDFFormer-B36}
\end{minipage}
\end{minipage}
\caption{Visualization of dynamic filter basis in the frequency domain on CDFFormers}\label{fig:cdfformer_filter}
\end{figure*}

Visualization of the dynamic filter basis in models which we do not mention in the main text, is shown in Figure~\ref{fig:dfformer_filter}, \ref{fig:cdfformer_filter}.
\subsection{More Models vs Human on Shape Bias}
A brief review of previous studies on shape bias shows that FFT-based token-mixers are expected to have more intense shape bias than not global token-mixers. In earlier studies, 
 \cite{baker2018deep}, CNNs are known to be insensitive to global shape information and instead demonstrate a strong reaction to a texture. Later, \cite{hermann2020origins} point out that several data augmentations reduce texture bias and \cite{tuli2021convolutional} that Transformers have more substantial shape bias than CNN. \cite{ding2022scaling} found that even CNNs can improve the shape bias due to a larger kernel size. Since the FFT-based model performs operations equivalent to those of a large kernel CNN, we expected that the FFT-based model would have a similar trend to CNNs with large kernels about shape bias, and we obtained such results on S18 scale models.

\begin{figure*}[tb]
\begin{minipage}[b]{0.49\hsize}
  \centering
  \includegraphics[height=1.\linewidth]{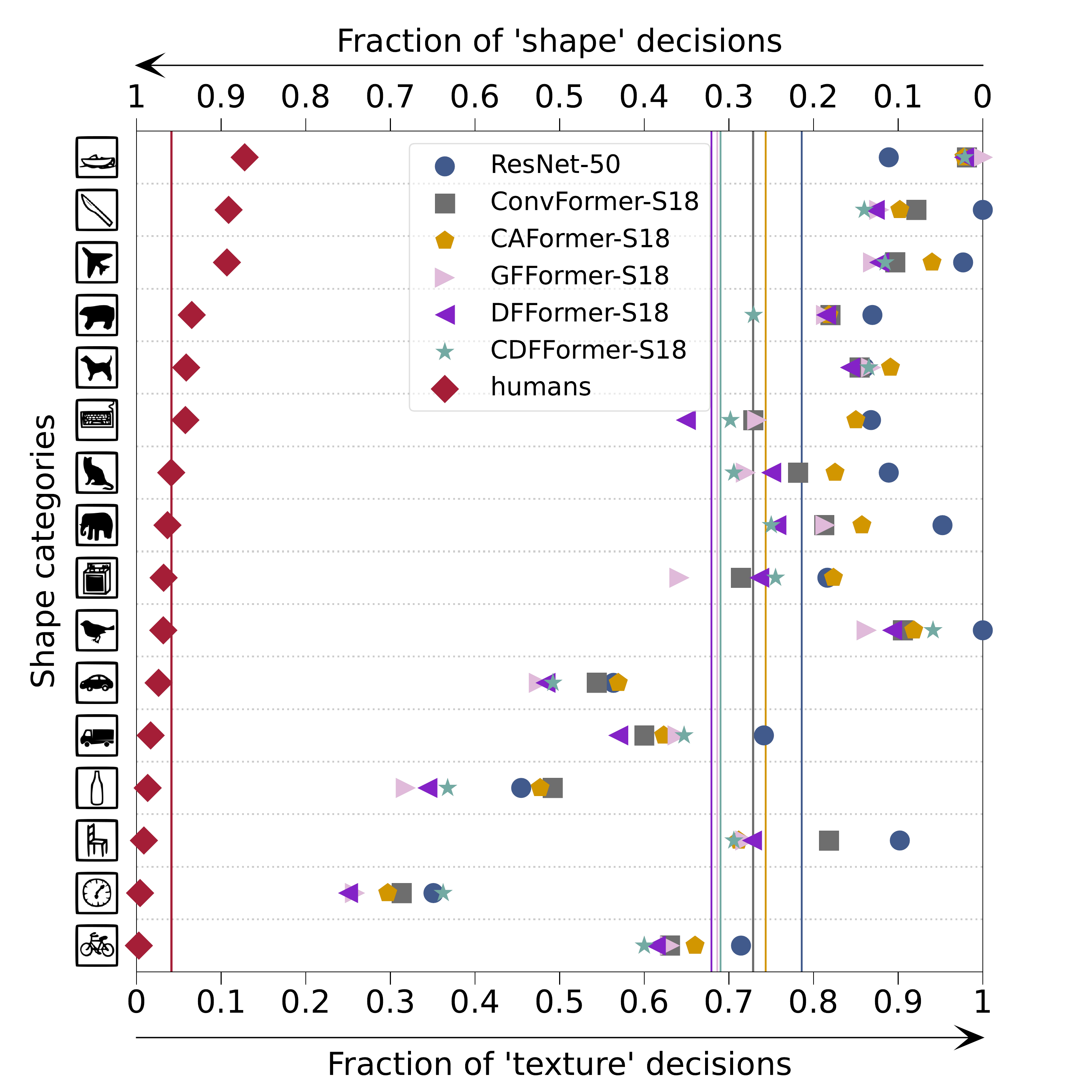}
  \subcaption{\label{fig:shape_bias_s18} S18 models vs human on shape bias}
\end{minipage}
\begin{minipage}[b]{0.49\hsize}
  \centering
  \includegraphics[height=1.\linewidth]{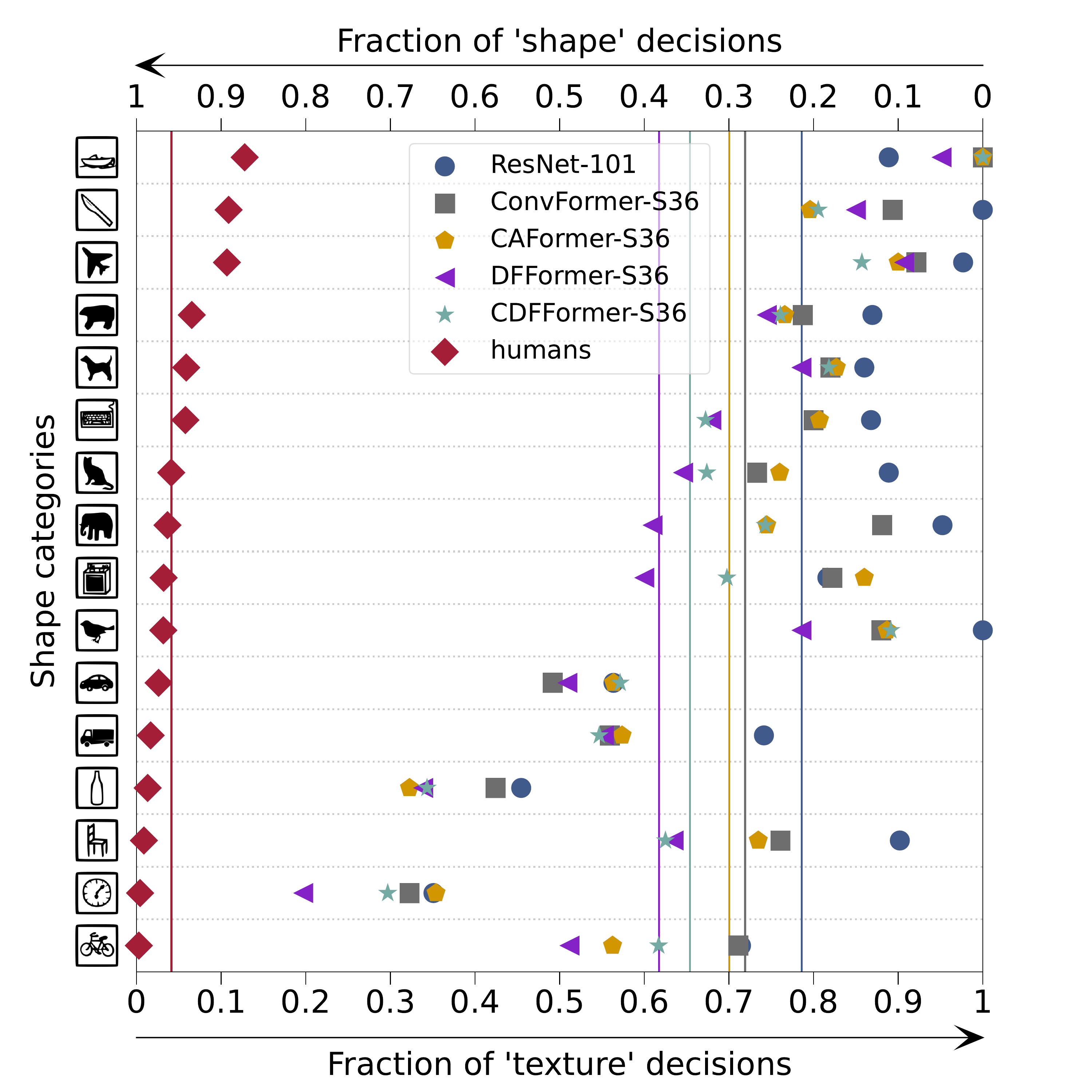}
  \subcaption{\label{fig:shape_bias_s36} S36 models vs human on shape bias}
\end{minipage}
\begin{minipage}[b]{0.49\hsize}
  \centering
  \includegraphics[height=1.\linewidth]{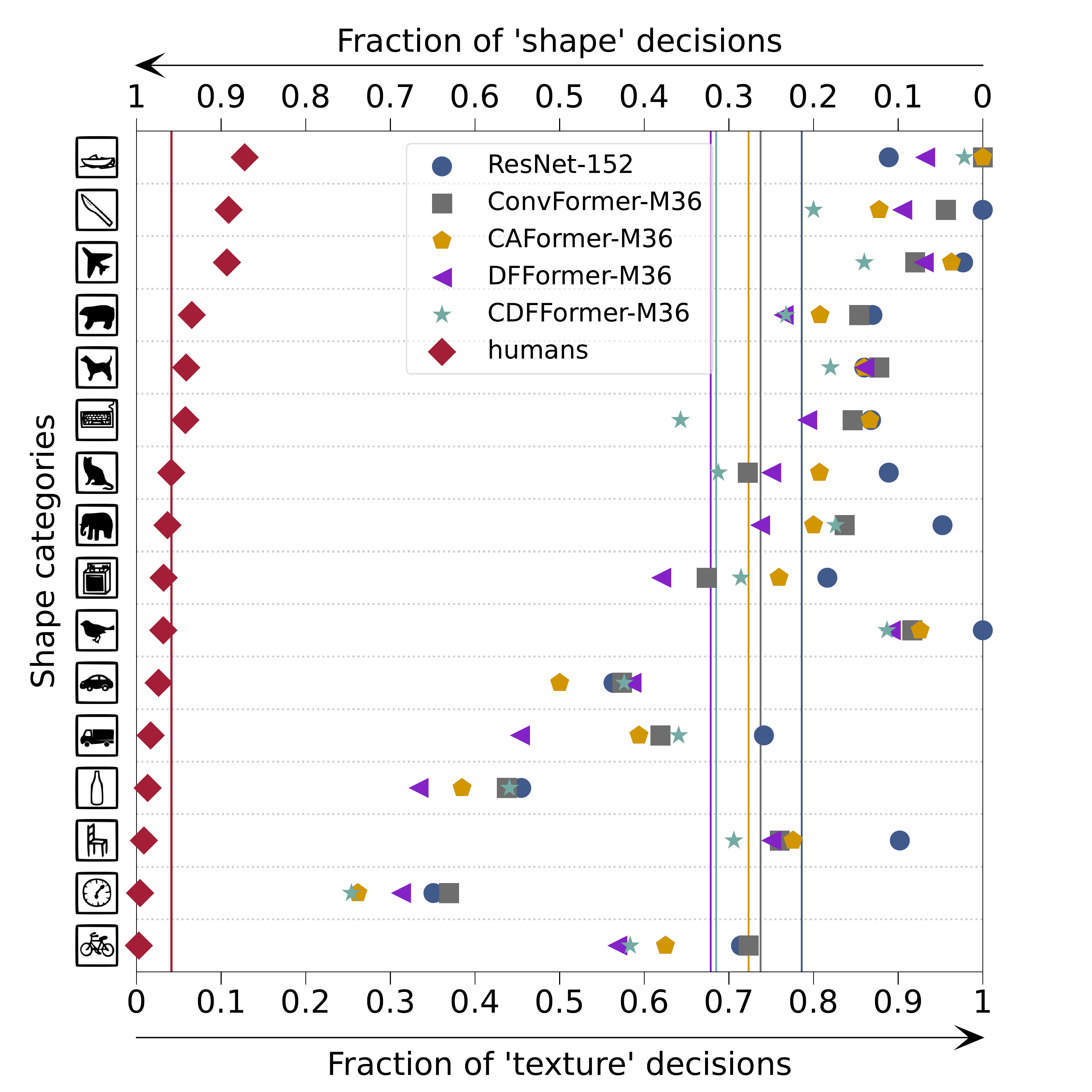}
  \subcaption{\label{fig:shape_bias_m36} M36 models vs human on shape bias}
\end{minipage}
\begin{minipage}[b]{0.49\hsize}
  \centering
  \includegraphics[height=1.\linewidth]{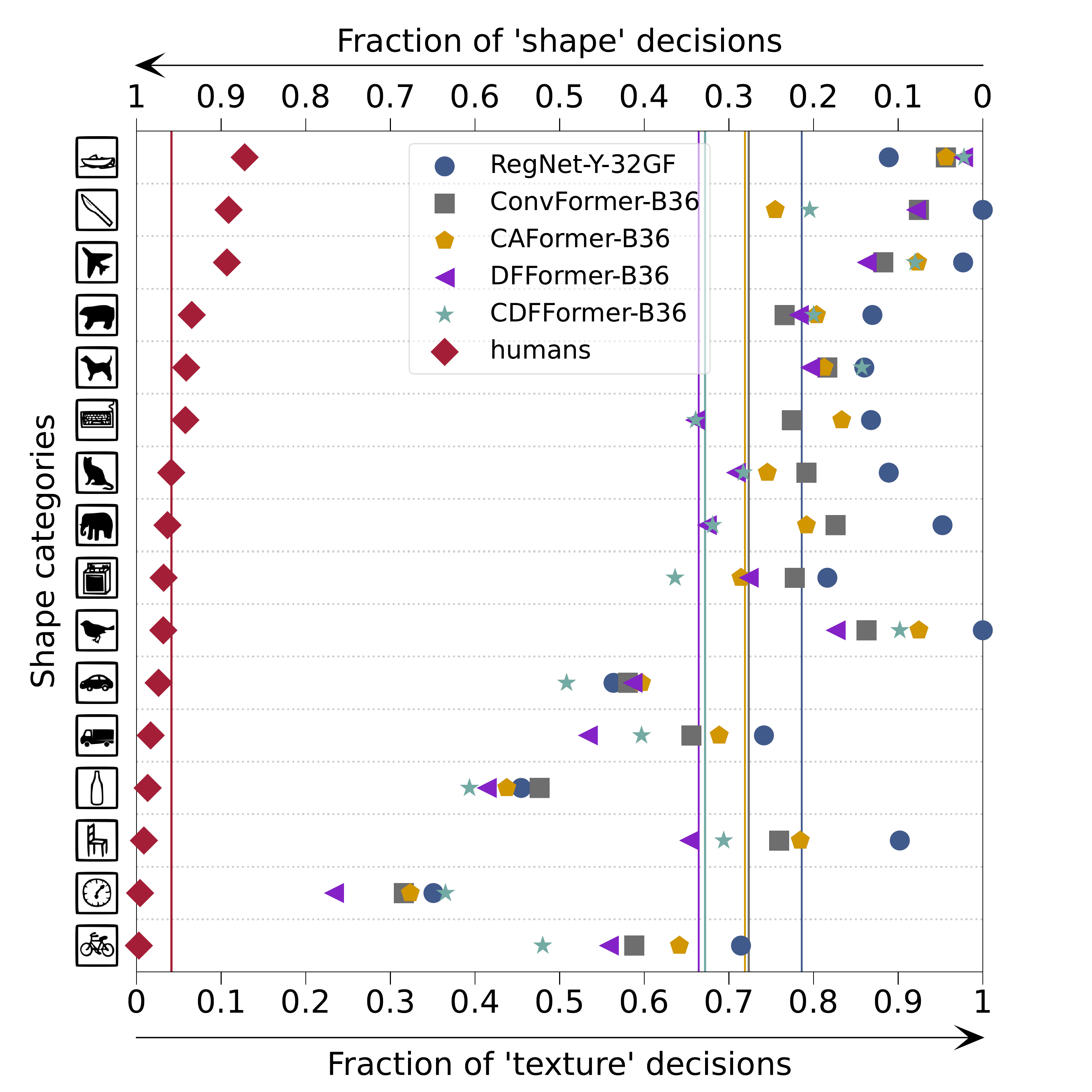}
  \subcaption{\label{fig:shape_bias_b36} B36 models vs human on shape bias}
\end{minipage}
  \caption{\label{fig:more_shape_bias} More  shape bias}
\end{figure*}

We already have discussed shape bias on S18 scale models, but similarly for other scales. Figure~\ref{fig:more_shape_bias} shows them of other scale models. The results tend to be the same as on S18 scale models.
\section{Detailed information}
\subsection{Code for Dynamic Filter}
The proposed dynamic filter can be implemented briefly by \texttt{PyTorch} \cite{paszke2019pytorch}. Algorithm~\ref{alg:dynamic_filter} shows the pseudocode for crucial parts of the dynamic filter.
\begin{algorithm*}[b]
\caption{Pseudocode for dynamic filter (PyTorch-like)}
\label{alg:dynamic_filter}
\definecolor{codeblue}{rgb}{0.25,0.5, 0.5}
\definecolor{codekw}{rgb}{0.85, 0.18, 0.50}
\lstset{
backgroundcolor=\color{white},
basicstyle=\fontsize{8pt}{8pt}\ttfamily\selectfont,
columns=fullflexible,
breaklines=true,
captionpos=b,
commentstyle=\fontsize{8pt}{8pt}\color{codeblue},
keywordstyle=\fontsize{8pt}{8pt}\color{codekw},
}
\begin{lstlisting}[language=python]
class DynamicFilter(nn.Module):
    def __init__(self, dim, expansion_ratio=2, bias=False, num_filters=4, size=14, **kwargs):
        super().__init__()
        size = to_2tuple(size)
        self.size = size[0]
        self.filter_size = size[1] // 2 + 1
        self.num_filters = num_filters
        self.dim = dim
        self.med_channels = int(expansion_ratio * dim)
        self.pwconv1 = nn.Linear(dim, self.med_channels, bias=bias)
        self.act = StarReLU()
        self.reweight = Mlp(dim, .25, num_filters * self.med_channels)
        self.complex_weights = nn.Parameter(
            torch.randn(self.size, self.filter_size, num_filters, 2,
                        dtype=torch.float32) * 0.02)
        self.pwconv2 = nn.Linear(self.med_channels, dim, bias=bias)

    def forward(self, x):
        B, H, W, _ = x.shape

        routeing = self.reweight(x.mean(dim=(1, 2))).view(B, self.num_filters, -1).softmax(dim=1)
        x = self.pwconv1(x)
        x = self.act(x)
        x = x.to(torch.float32)
        x = torch.fft.rfft2(x, dim=(1, 2), norm='ortho')
        complex_weights = torch.view_as_complex(self.complex_weights)
        routeing = routeing.to(torch.complex64)
        weight = torch.einsum('bfc,hwf->bhwc', routeing, complex_weights)
        weight = weight.view(-1, self.size, self.filter_size, self.med_channels)
        x = x * weight
        x = torch.fft.irfft2(x, s=(H, W), dim=(1, 2), norm='ortho')
        x = self.pwconv2(x)
        return x


class Mlp(nn.Module):
    def __init__(self, dim, mlp_ratio, out_features=None, bias=False, **kwargs):
        super().__init__()
        in_features = dim
        out_features = out_features or in_features
        hidden_features = int(mlp_ratio * in_features)

        self.fc1 = nn.Linear(in_features, hidden_features, bias=bias)
        self.act = StarReLU()
        self.fc2 = nn.Linear(hidden_features, out_features, bias=bias)

    def forward(self, x):
        x = self.fc1(x)
        x = self.act(x)
        x = self.fc2(x)
        return x
\end{lstlisting}

\end{algorithm*}
\subsection{Code for Visualization}
We utilized Algorithm~\ref{alg:visualization} to visualize the complex parameters on the frequency domain in the global filter and dynamic filter.
\begin{algorithm*}[b]
\caption{Pseudocode of visualization of weights in global and dynamic filters (PyTorch-like)}
\label{alg:visualization}
\definecolor{codeblue}{rgb}{0.25, 0.5, 0.5}
\definecolor{codekw}{rgb}{0.85, 0.18, 0.50}
\lstset{
backgroundcolor=\color{white},
basicstyle=\fontsize{8pt}{8pt}\ttfamily\selectfont,
columns=fullflexible,
breaklines=true,
captionpos=b,
commentstyle=\fontsize{8pt}{8pt}\color{codeblue},
keywordstyle=\fontsize{8pt}{8pt}\color{codekw},
}
\begin{lstlisting}[language=python]
def visualize(weight):
    # weight.shape: (height, width, number_of_filter, 2)
    # weight.dtype: torch.float
    weight = torch.view_as_complex(weight)
    amplitude = weight.abs() + 1e-6
    amplitude = amplitude.log()
    amplitude = amplitude.sigmoid()
    h, w, _ = amplitude.shape
    if h % 2 == 0:
        amplitude = torch.cat([amplitude, torch.flip(amplitude[:, 1:-1], dims=[1])], dim=1)
    else:
        amplitude = torch.cat([amplitude, torch.flip(amplitude[:, 1:], dims=[1])], dim=1)
    h, w, _ = amplitude.shape
    amplitude = torch.roll(amplitude, shifts=(int(h / 2), int(w / 2)), dims=(0, 1))
    return amplitude
\end{lstlisting}

\end{algorithm*}
\subsection{ImageNet-1K Training Setting}
We provide ImageNet-1K training settings in Table~\ref{tab:hyperparameters}. These settings are used in the main results.
\begin{table*}[b]
    \centering
\begin{NiceTabular}{lc}
\toprule
Training config. & {\makecell[c]{DFFormer- \& CDFFormer- \\ S18/S36/M36/B36}} \\
\hline
dataset & ImageNet-1K~\cite{krizhevsky2012imagenet} \\
resolution & $224^2$ \\
optimizer & AdamW~\cite{loshchilov2017decoupled} \\
base learning rate & 1e-3 \\
weight decay & 0.05 \\
optimizer $\epsilon$ & 1e-8 \\
optimizer momentum & $\beta_1,=0.9, \beta_2{=}0.999$ \\
batch size & 1024 \\
training epochs & 300 \\
learning rate schedule & cosine decay \\
lower learning rate bound & 1e-6 \\
warmup epochs & 20 \\
warmup schedule & linear \\
warmup learning rate & 1e-6 \\
cooldown epochs & 10 \\
crop ratio & 1.0 \\
RandAugment~\cite{cubuk2020randaugment} & (9, 0.5) \\
Mixup $\alpha$~\cite{zhang2017mixup} & 0.8 \\
CutMix $\alpha$~\cite{yun2019cutmix} & 1.0 \\
random erasing~\cite{zhong2020random} & 0.25  \\
label smoothing~\cite{szegedy2016rethinking} & 0.1  \\
stochastic depth~\cite{huang2016deep} & 0.2/0.3/0.4/0.6 \\
LayerScale~\cite{touvron2021going} init. & None \\
ResScale~\cite{shleifer2021normformer} init. & 1.0 (only for the last two stages) \\
\bottomrule
\end{NiceTabular}
    \caption{\label{tab:hyperparameters}
    Hyper-parameters of classification on ImageNet-1K}
\end{table*}
\subsection{Replacement of Filters in Ablation}
We replaced dynamic filters with global filters and AFNOs in the ablation study. Here, we describe how to replace them. When replaced by global filters, $\mathcal{L}$ is defined as follows:
\begin{equation}
\mathcal{L}(\cdot)=\mathcal{G}(\cdot).    
\end{equation}
If replaced by AFNOs,
\begin{equation}
\mathcal{L}(\cdot)=\AFNO(\cdot).    
\end{equation}
\subsection{Detail of Advantages at Higher Resolutions}
We discussed the throughput and peak memory changes during inference when varying the resolution. In the main text, the original measurements of the plotted results are listed in Table~\ref{tab:throughput_memory}.
\begin{table*}[b]
    \centering
    \begin{tabular}{ll|rrrr|rrrr}
\toprule
&&\multicolumn{4}{c}{Throughput(images/s)} & \multicolumn{4}{c}{Peak memory (MB)} \\
Size&Model\textbackslash Resolution&$256^2$&$512^2$&$768^2$&$1024^2$&$256^2$&$512^2$&$768^2$&$1024^2$ \\ \hline
\multirow{4}{*}{S18} &ConvFormer &199.88 &149.70 &73.25 &42.26 &171 &371 &707 &1175 \\
&CAFormer &202.64 &113.31 &35.91 &15.61 &169 &407 &1435 &4131 \\
&DFFormer &148.59 &111.81 &54.45 &30.54 &183 &385 &723 &1192 \\
&CDFFormer &168.99 &129.86 &63.60 &36.97 &183 &384 &720 &1188 \\ \hline
\multirow{4}{*}{S36} &ConvFormer &107.43 &78.95 &38.66 &22.51 &232 &421 &757 &1225 \\
&CAFormer &109.14 &59.07 &18.53 &8.08 &229 &481 &1509 &4205 \\
&DFFormer &76.18 &58.46 &28.55 &16.39 &265 &444 &782 &1254 \\
&CDFFormer &90.67 &68.48 &33.73 &19.70 &263 &441 &778 &1246 \\ \hline
\multirow{4}{*}{M36} &ConvFormer &115.30 &58.67 &28.31 &16.33 &351 &620 &1116 &1808 \\
&CAFormer &108.48 &45.38 &14.53 &6.43 &340 &625 &1856 &5088 \\
&DFFormer &77.51 &43.67 &20.89 &11.69 &397 &653 &1151 &1846 \\
&CDFFormer &87.58 &51.87 &24.95 &14.58 &393 &648 &1145 &1837 \\ \hline
\multirow{4}{*}{B36} &ConvFormer &111.39 &43.84 &21.16 &12.13 &577 &907 &1563 &2479 \\
&CAFormer &106.96 &33.60 &10.89 &4.79 &583 &960 &2592 &6895 \\
&DFFormer &77.90 &33.03 &15.85 &8.71 &663 &966 &1624 &2545 \\
&CDFFormer &86.22 &39.03 &18.93 &10.89 &655 &959 &1615 &2531 \\
\bottomrule
\end{tabular}
\caption{\label{tab:throughput_memory}\textbf{Throughput vs. resolution and peak memory vs. resolution} these have been benchmarked on a V100 with 16GB 
memory at a batch size of four. In this paper, we propose the models highlighted with \colorbox{bgpink}{pink} as shown are proposed in this paper.}
\end{table*}

\end{document}